\documentclass[format=acmsmall, review=false, screen=true]{acmart}

\usepackage{graphicx}
\usepackage{graphicx,calc}

\newlength\myheight
\newlength\mydepth
\settototalheight\myheight{Xygp}
\newcommand*\inlinegraphics[1]{%
  \settototalheight\myheight{Xygp}%
  \settodepth\mydepth{Xygp}%
  \raisebox{-\mydepth}{\includegraphics[height=\myheight]{#1}}%
}

\begin{document}

\title{Histograms of Points, Orientations, and Dynamics of Orientations Features for Hindi Online Handwritten Character Recognition}

\author{Anand Sharma}
\email{anand.sharma@miet.ac.in}
\affiliation{
\institution{MIET}
\city{Meerut}
\country{India}}
\author{A. G. Ramakrishnan}
\email{agr@iisc.ac.in}
\affiliation{
\institution{Indian Institute of Science}
\city{Bangaluru}
\country{India}}

\begin{abstract}
\indent A set of features independent of character stroke direction and order variations is proposed for online handwritten character recognition. A method is developed that maps features like co-ordinates of points, orientations of strokes at points, and dynamics of orientations of strokes at points spatially as a function of co-ordinate values of the points and computes histograms of these features from different regions in the spatial map. \\
\indent Different features like spatio-temporal, discrete Fourier transform, discrete cosine transform, discrete wavelet transform, spatial, and histograms of oriented gradients used in other studies for training classifiers for character recognition are considered. The classifier chosen for classification performance comparison, when trained with different features, is support vector machines (SVM).\\
\indent The character datasets used for training and testing the classifiers consist of online handwritten samples of 96 different Hindi characters. There are 12832 and 2821 samples in training and testing datasets, respectively.\\
\indent SVM classifiers trained with the proposed features has the highest classification accuracy of 92.9\% when compared to the performances of SVM classifiers trained with the other features and tested on the same testing dataset. Therefore, the proposed features have better character discriminative capability than the other features considered for comparison. 
\end{abstract}



\maketitle

\section{Introduction}
An online handwritten character is written as a sequence of strokes. A stroke is a sequence of points produced by movement of a pen on a touch sensitive screen between ped-down state and pen-up state. When the pen is touching the screen, it is in a pen-down state; and when it is not touching the screen, in a pen-up state. Samples of online handwritten characters written by different individuals have lot of variations because different individuals have different writing styles. A set of features is effective for recognition of such online characters if the features have large variability for the samples of characters belonging to different character classes compared to those belonging to the same class Tappert et al. \cite{troaa}. Such features facilitate design of high accuracy character classifiers.\\
\indent Different studies have used different types of features for increasing the recognition accuracy of classifiesr. Both spatio-temporal (ST) and discrete Fourier transform (DFT) features are used by Swethalakshmi \cite{ftrmm} for training stroke based classifiers. Sundaram et al. \cite{troajj} use ST features for training character classifiers as part of word recognition algorithm. Discrete wavelet transform (DWT) features are used for character recognition by Kunte et al. \cite{ftri} and discrete cosine transform (DCT) features are used for character recognition by Kubatur et al. \cite{troav}. Bhattacharya et al. \cite{ftrh} use direction code based features for character recognition. Bahlmann \cite{ftra} combines linear and directional features into a single multivariate probability density function and estimates parameters of approximation to this density function for recognition of characters. Six dimensional features at dominant points in strokes of characters are used by Ma et al. \cite{ftrj} to train continuous density hidden Markov model (HMM) for character recognition. A combination of online and offline features is used by connell et al. \cite{troan}, Okamoto et al. \cite{ftrc}, Okamoto et al. \cite{ftrk} and Jaeger et al. \cite{ftrl}. Both the online features and the combination of online and offline features used in the above studies vary with variations in order and direction of strokes in the characters.\\
\indent Studies done by Kawamura et al. \cite{ftrg} and  Hamanaka et al. \cite{ftre} have used offline features extracted from online handwritten characters for training classifiers for character recognition. Mehrotra et al. \cite{troax} use spatial (SP) features to train convolutional neural network classifier. Belhe et al. \cite{troaw} use histograms of oriented gradients (HOG) features to train HMM classifier. These features are independent of variations in order and direction of strokes in characters.\\
\indent Features used in different studies on Hindi online handwritten character recognition have been extracted from different datasets and from character sets of different sizes. Comparison of character classifier performance on these different features is therefore not possible. Some of the features from the literature that have been considered for this study are ST, DFT, DCT, DWT, SP, and HOG features and have been considered so that classifier performance on the designed features can be compared with those on the considered features extracted from the same dataset.\\
\indent Histograms of points, orientations, and dynamics of orientations (HPOD) features is developed in this study. HPOD feature extraction method spatially maps different features and computes histograms of these features from different overlapping regions of the spatial map. These features are independent of stroke order and direction variations. HPOD features are new and have been used for Hindi online handwritten character recognition for the first time to the best of our knowledge.\\
\indent This paper is organized into five Sections. Section 2 describes Hindi character set considered for recognition and the corresponding character dataset. Different feature extraction methods along with the proposed feature extraction method are explained in Section 3. Experiments and results are given in Section 4. Section 5 summarizes the contribution of this work. 

\section{Hindi character} 
Hindi language is written using Devanagari script \cite{troiscii}. Hindi character set is a subset of Devanagari character set. Online handwritten samples of most of the basic Hindi characters are used for the study of feature design and extraction and comparison of classifiers' performance trained with the extracted features.\\
\subsection{Hindi character set}
\indent Hindi characters considered in our work are vowels, consonants, half consonants, nasalization sign, vowel omission sign, vowel signs, consonant with vowel sign, conjuncts, and consonant clusters. Strokes with different shapes are also considered for recognition in addition to these characters. Figure 1 shows the subset of Hindi vowels considered for recognition. The other vowels can be obtained by a combination of these vowels and one or more vowel signs. For example, the vowel
 
 {\inlinegraphics{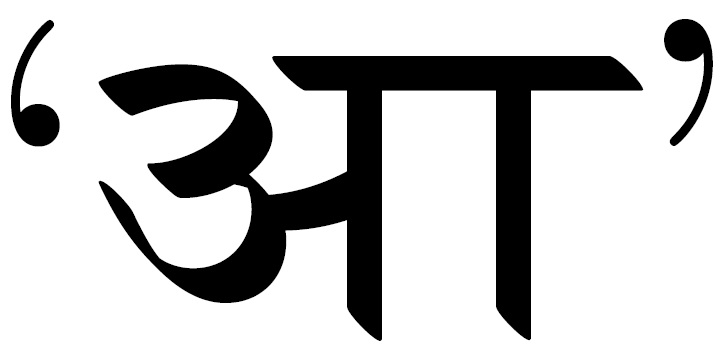}} can be obtained as a combination of symbol 1 in Fig. 1 and symbol 67 in Fig. 4.  
\begin{figure}[ht]
\begin{center}
$\begin{array}{cccccc}
\includegraphics[width=0.4 in]{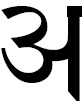}&
\includegraphics[width=0.4 in]{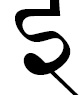}&
\includegraphics[width=0.4 in]{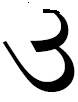}&
\includegraphics[width=0.5 in]{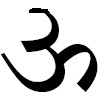}&
\includegraphics[width=0.6 in]{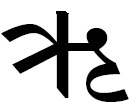}&
\includegraphics[width=0.35 in]{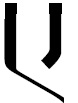}\\
1&2&3&4&5&6\\
\end{array}$
\caption{Basic Hindi vowels used as recognition primitives.}
\label{vowel}
\end{center}
\end{figure}
Figure 2 shows the Hindi consonants considered for recognition. All the consonants have an implicit vowel `a' \cite{troiscii}.
\begin{figure}[ht]
\begin{center}
$\begin{array}{cccccccccc}
\includegraphics[width=0.45 in]{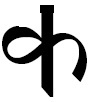}&
\includegraphics[width=0.45 in]{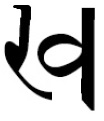}&
\includegraphics[width=0.4 in]{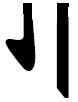}&
\includegraphics[width=0.4 in]{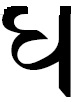}&
\includegraphics[width=0.4 in]{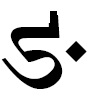}&
\includegraphics[width=0.4 in]{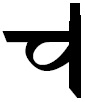}&
\includegraphics[width=0.4 in]{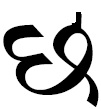}&
\includegraphics[width=0.4 in]{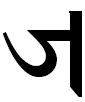}&
\includegraphics[width=0.4 in]{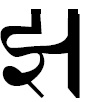}&
\includegraphics[width=0.4 in]{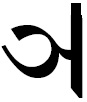}\\
7&8&9&10&11&12&13&14&15&16\\
\includegraphics[width=0.4 in]{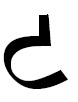}&
\includegraphics[width=0.4 in]{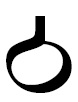}&
\includegraphics[width=0.4 in]{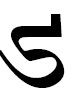}&
\includegraphics[width=0.4 in]{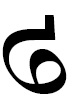}&
\includegraphics[width=0.4 in]{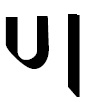}&
\includegraphics[width=0.4 in]{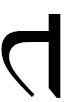}&
\includegraphics[width=0.4 in]{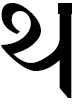}&
\includegraphics[width=0.4 in]{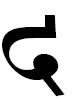}&
\includegraphics[width=0.4 in]{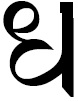}&
\includegraphics[width=0.4 in]{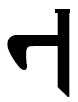}\\
17&18&19&20&21&22&23&24&25&26\\
\includegraphics[width=0.4 in]{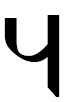}&
\includegraphics[width=0.4 in]{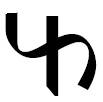}&
\includegraphics[width=0.4 in]{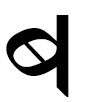}&
\includegraphics[width=0.4 in]{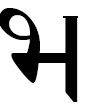}&
\includegraphics[width=0.4 in]{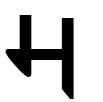}&
\includegraphics[width=0.4 in]{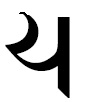}&
\includegraphics[width=0.4 in]{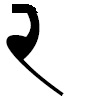}&
\includegraphics[width=0.4 in]{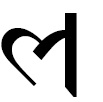}&
\includegraphics[width=0.4 in]{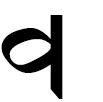}&
\includegraphics[width=0.4 in]{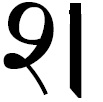}\\
27&28&29&30&31&32&33&34&35&36\\
\includegraphics[width=0.4 in]{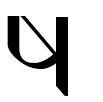}&
\includegraphics[width=0.4 in]{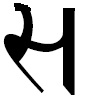}&
\includegraphics[width=0.4 in]{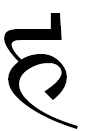}\\
37&38&39&&&&&&&\\
\end{array}$
\caption{Hindi consonants included as part of the recognized classes.}
\label{consonant}
\end{center}
\end{figure}
Half consonants are the corresponding consonants with their implicit vowel muted. Figure 3 shows Hindi half consonants. The nasalization sign indicates nasalization of the character the sign is written over. The vowel omission sign is used to mute the implicit vowel of the consonants. The vowel sign is used to modify the implicit vowel of a consonant. Figure 4(65) shows the nasalization sign, Figure 4(66) shows the vowel omission sign, Figures 4(67)-(72) show vowel signs and Figure 4(73) shows a consonant modified by vowel sign.
\begin{figure}[ht]
\begin{center}
$\begin{array}{cccccccccc}
\includegraphics[width=0.45 in]{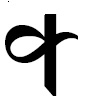}&
\includegraphics[width=0.45 in]{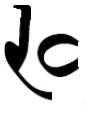}&
\includegraphics[width=0.4 in]{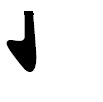}&
\includegraphics[width=0.4 in]{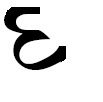}&
\includegraphics[width=0.4 in]{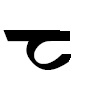}&
\includegraphics[width=0.4 in]{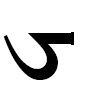}&
\includegraphics[width=0.4 in]{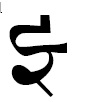}&
\includegraphics[width=0.4 in]{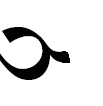}&
\includegraphics[width=0.4 in]{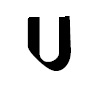}&
\includegraphics[width=0.4 in]{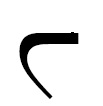}\\
40&41&42&43&44&45&46&47&48&49\\
\includegraphics[width=0.4 in]{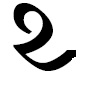}&
\includegraphics[width=0.4 in]{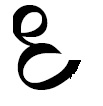}&
\includegraphics[width=0.4 in]{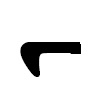}&
\includegraphics[width=0.4 in]{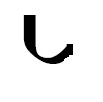}&
\includegraphics[width=0.4 in]{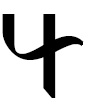}&
\includegraphics[width=0.4 in]{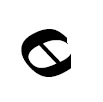}&
\includegraphics[width=0.4 in]{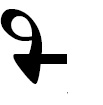}&
\includegraphics[width=0.4 in]{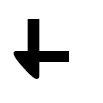}&
\includegraphics[width=0.4 in]{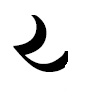}&
\includegraphics[width=0.4 in]{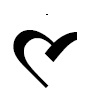}\\
50&51&52&53&54&55&56&57&58&59\\
\includegraphics[width=0.4 in]{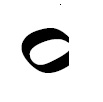}&
\includegraphics[width=0.4 in]{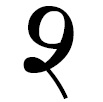}&
\includegraphics[width=0.4 in]{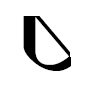}&
\includegraphics[width=0.4 in]{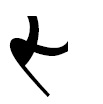}&
\includegraphics[width=0.4 in]{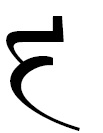}&\\
60&61&62&63&64&&&&&\\
\end{array}$
\caption{Hindi half consonants considered for recognition.}
\label{hconsonant}
\end{center}
\end{figure}
  
\begin{figure}[ht!]
\begin{center}
$\begin{array}{ccccccccc}
\includegraphics[width=0.45 in]{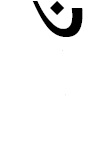}&
\includegraphics[width=0.45 in]{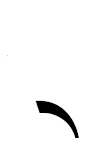}&
\includegraphics[width=0.4 in]{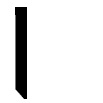}&
\includegraphics[width=0.4 in]{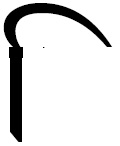}&
\includegraphics[width=0.4 in]{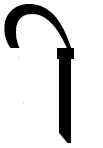}&
\includegraphics[width=0.4 in]{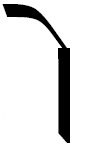}&
\includegraphics[width=0.4 in]{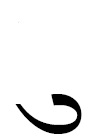}&
\includegraphics[width=0.4 in]{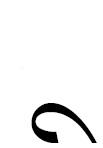}&
\includegraphics[width=0.4 in]{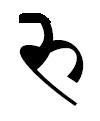}\\
65&66&67&68&69&70&71&72&73\\
\end{array}$
\caption{Some signs and consonant with vowel sign considered for recognition. (65) Nasalization sign. (66) Vowel omission sign. (67)-(72) Vowel signs. (73) Consonant with vowel sign.}
\label{nasal}
\end{center}
\end{figure}
Conjuncts are cluster of consonants, where the implicit vowels of all but the last consonant of the cluster are muted. There are cluster of consonants where implicit vowel of last consonant is muted. The conjuncts and the consonant clusters considered for recognition are shown in Fig. 5.
\begin{figure}[ht!]
\begin{center}
$\begin{array}{cccccccccc}
\includegraphics[width=0.4 in]{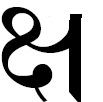}&
\includegraphics[width=0.4 in]{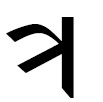}&
\includegraphics[width=0.4 in]{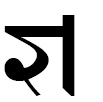}&
\includegraphics[width=0.4 in]{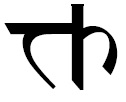}&
\includegraphics[width=0.4 in]{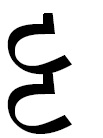}&
\includegraphics[width=0.4 in]{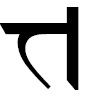}&
\includegraphics[width=0.4 in]{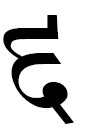}&
\includegraphics[width=0.4 in]{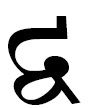}&
\includegraphics[width=0.4 in]{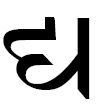}&
\includegraphics[width=0.4 in]{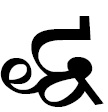}\\
74&75&76&77&78&79&80&81&82&83\\
\includegraphics[width=0.4 in]{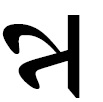}&
\includegraphics[width=0.4 in]{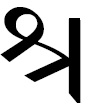}&
\includegraphics[width=0.4 in]{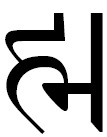}&
\includegraphics[width=0.4 in]{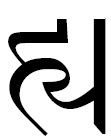}&
\includegraphics[width=0.4 in]{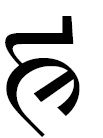}&
\includegraphics[width=0.4 in]{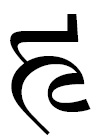}&
\includegraphics[width=0.4 in]{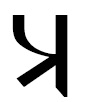}&
\includegraphics[width=0.4 in]{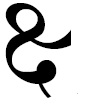}&
\includegraphics[width=0.4 in]{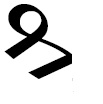}&
\includegraphics[width=0.4 in]{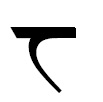}\\
84&85&86&87&88&89&90&91&92&93\\
\end{array}$
\caption{Conjuncts and consonant clusters included as part of recognized classes. Conjuncts are cluster of consonants, where only implicit vowel of the last consonant is not muted. Symbols, where implicit vowel of the last consonant of a consonant cluster is muted, are also used. (74)-(90) Frequently used Hindi conjuncts. (91)-(93) Hindi consonant clusters with implicit vowel of the last consonant muted.}
\label{conjunct}
\end{center}
\end{figure}
The other strokes considered for recognition are shown in Fig. 6.  
\begin{figure}[ht]
\begin{center}
$\begin{array}{ccc}
\includegraphics[width=0.45 in]{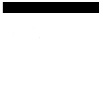}&
\includegraphics[width=0.45 in]{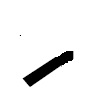}&
\includegraphics[width=0.4 in]{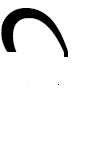}\\
94&95&96\\
\end{array}$
\caption{Strokes with different shapes considered for recognition.}
\label{stkos}
\end{center}
\end{figure}
\subsection{Hindi character dataset}
\indent The databases of Hindi online handwritten characters used for this study have been collected from Center for Development of Advanced Computing (CDAC) and Hewlett Packard (HP) datasets \cite{trohp}. Only those characters that are not cursive and have been written correctly are used. Figures 7(a) and 7(b) show samples of Hindi online handwritten characters obtained from the dataset used. Each of these characters is a sequence of strokes written in a particular order. Each stroke is written in a specific sequence of directions. The header lines over all the characters have been removed because they do not contribute to the structure of the characters.\\
\indent The total number of character samples in the dataset is 15653. The dataset is divided to form the training and testing datasets. The training set consists of 12832 samples and the test set, 2821 samples. These datasets consist of samples from 96 different character classes, with an average of 133 samples per class in the training set and 29, in the testing set.
\begin{figure}[ht]
\begin{center}
$\begin{array}{cc}
\includegraphics[width=2.5 in]{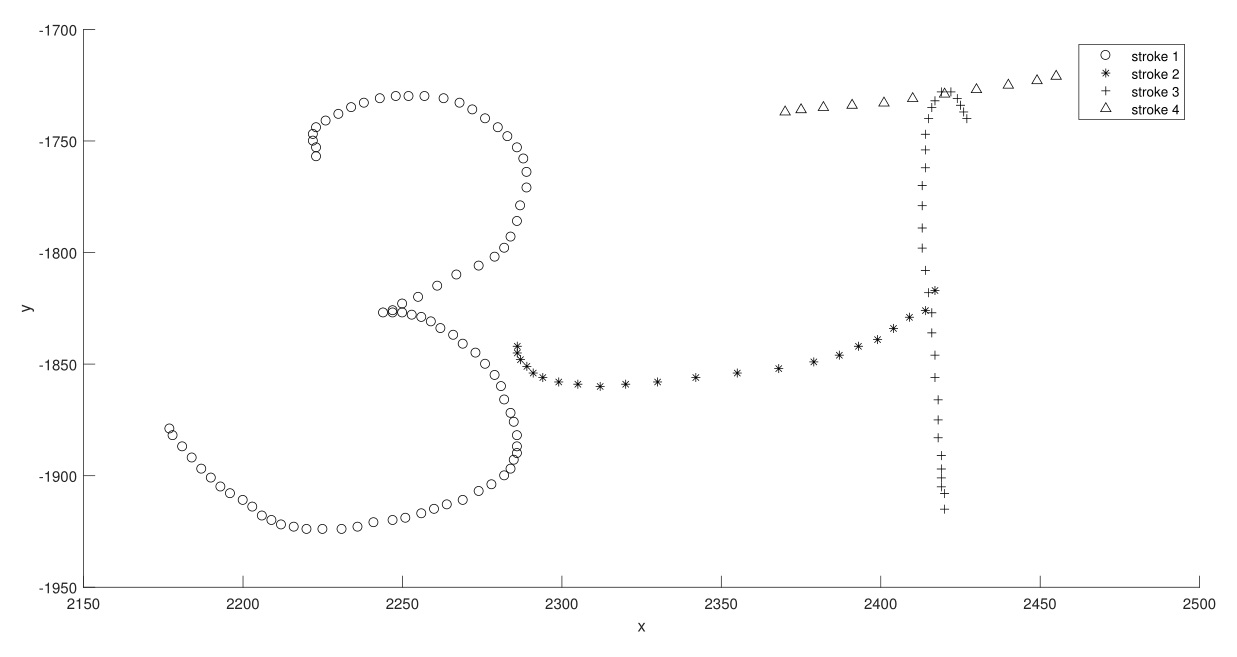}&
\includegraphics[width=2.5 in]{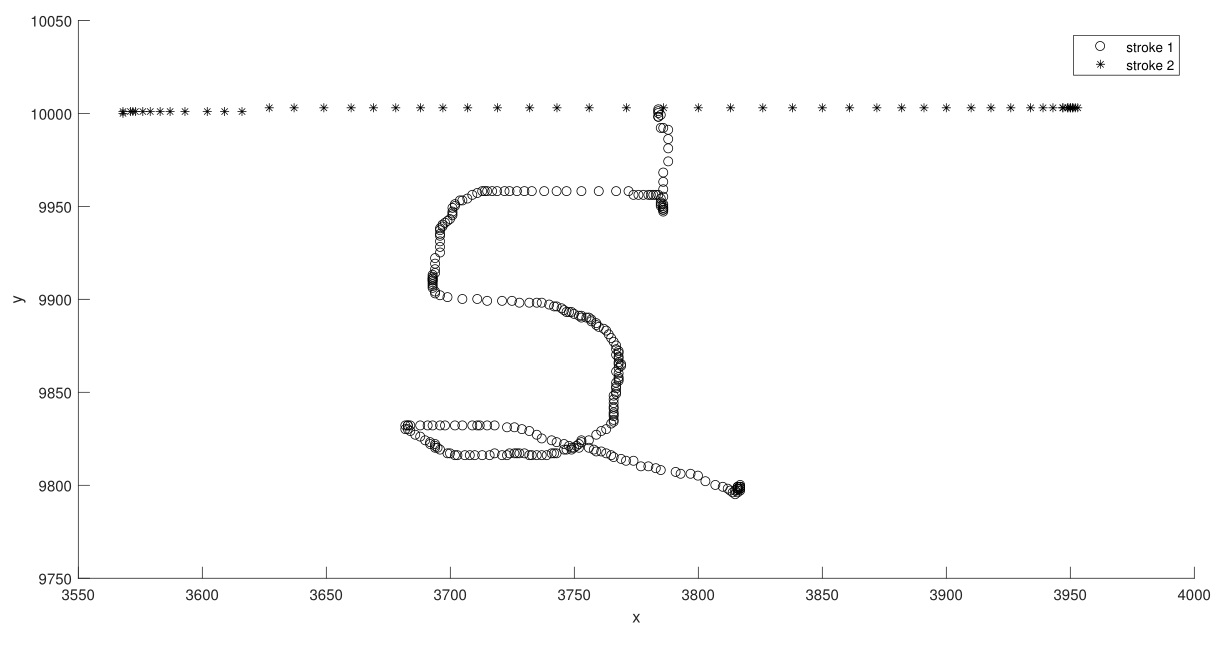}\\
(a)&(b)\\
\end{array}$
\end{center}
\caption{Samples of Hindi online handwritten characters. (a) Online character produced using four strokes. Stroke1, stroke2, and stroke3 correspond to structure of the character. Stroke4 is a header line and does not conribute to the structure of the character. (b) Online character produced using two strokes. Stroke1 corresponds to structure of the character. Stroke2 is a header line.}
\label{ohwr}
\end{figure}
\subsection{Preprocessing of samples in the dataset}
\indent The samples of online handwritten characters in the constructed datasets have got variations because they have been produced by different individuals at different times. Variations among the samples of the characters can be classified as external variations and internal variations. External variations are variations that can be removed by preprocessing or feature design. Internal variations are caused by deviation of handwritten character structure from the corresponding ideal character structure and are very difficult to remove. It is important to preprocess the characters so that feature extraction from the characters can be done in a uniform fashion.\\
\indent Some of the external variations removed from handwritten characters in the datasets are as follows. Repeated points are consecutive points in the character that have the same co-ordinate values and do not contribute to the structure of the character and so are detected and removed. Both the x- and y-co-ordinate values of the points in the characters are mapped to the range $[0\,\,1]$ by linear transformation. This removes variations in location and size of the characters. Distance between all the consecutive points in a character are made equal to a constant $\Delta$, thus removing variation in speed at which the character has been written. The value of $\Delta$ depends on the length of the character along its strokes for ST, DFT, DCT, and DWT features. The value of $\Delta$ can be chosen indepenently for SP, HOG and HPOD features. A character stroke trace roughness is removed by linear filtering the sequences of x- and y-co-ordinates of points in the stroke.

\subsection{Preprocessed character dataset}
\indent After preprocessing, an online handwritten character is represented as a sequence of strokes $C=(S_1,\dots,S_{N_S}),$ where $N_S$ is the number of strokes in the character. A stroke is represented as
\begin{equation}\label{stroke}S_i=[{p_1^{S_i}}^T;\dots ;{p_n^{S_i}}^T;\dots ;{p_{N_{S_i}}^{S_i}}^T],\,\, S_i\in[0\,\, 1]^{N_{S_i}\times2},\,\, 1\le i\le N_S,\,\,p_n^{S_i}=[x_n,\,\, y_n]^T,\,\, p_n^{S_i}\in[0 \,\,1]^{2\times1}.\end{equation}
$[x,\,y]$ is row-wise concatenation of $x$ and $y$, $[x;\,y]$ is column-wise concatenation of $x$ and $y$, and $T$ is transpose of a matrix. Point $p_n^{S_i}$ is the $n^{th}$ point in the stroke $S_i$ and $N_{S_i}$ is the number of points in the stroke $S_i$. Figure 8 shows samples of online handwritten characters before and after preprocessing.  
\begin{figure}[ht]
\begin{center}
$\begin{array}{cc}
\includegraphics[width=2 in]{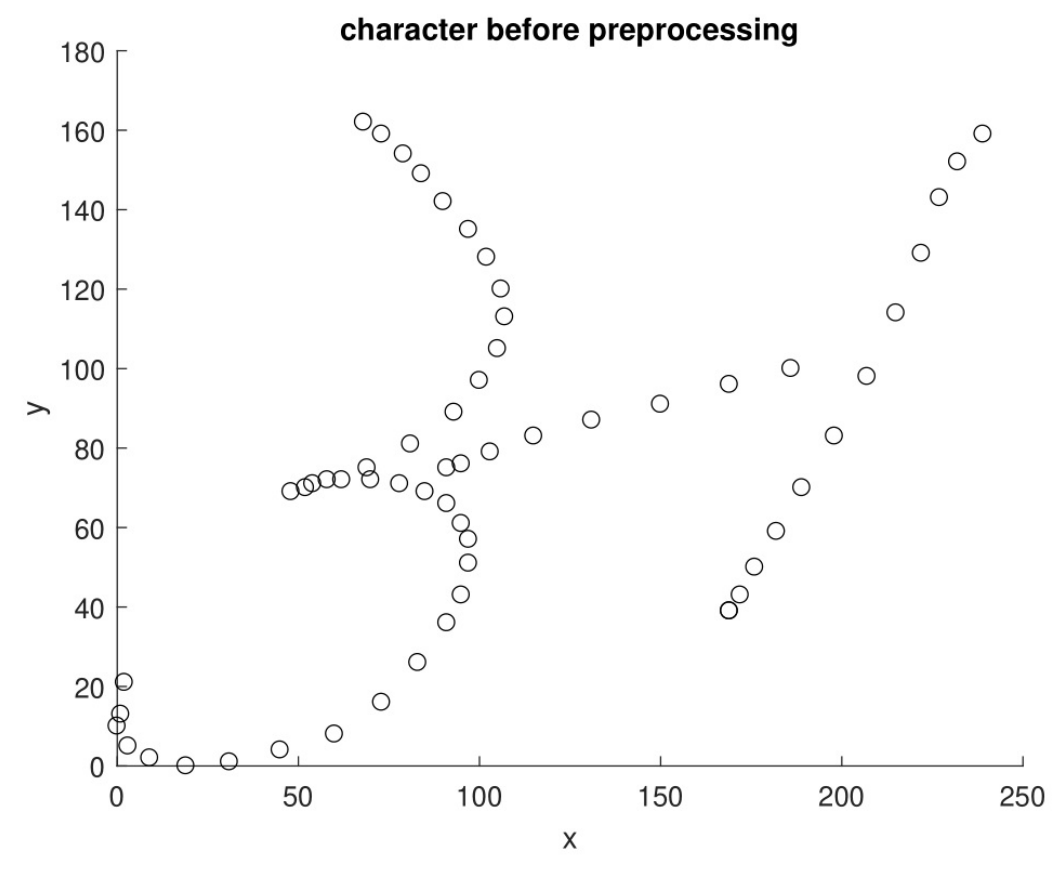}&
\includegraphics[width=2 in]{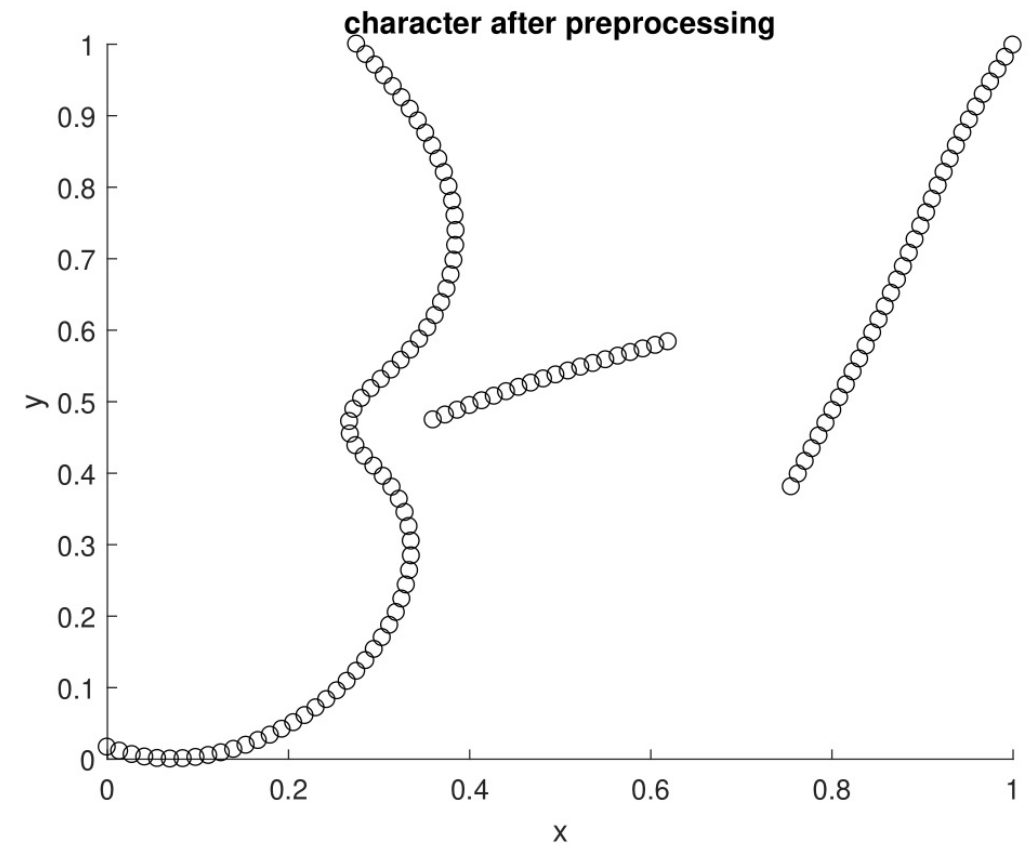}\\
(a)&(b)\\
\includegraphics[width=2 in]{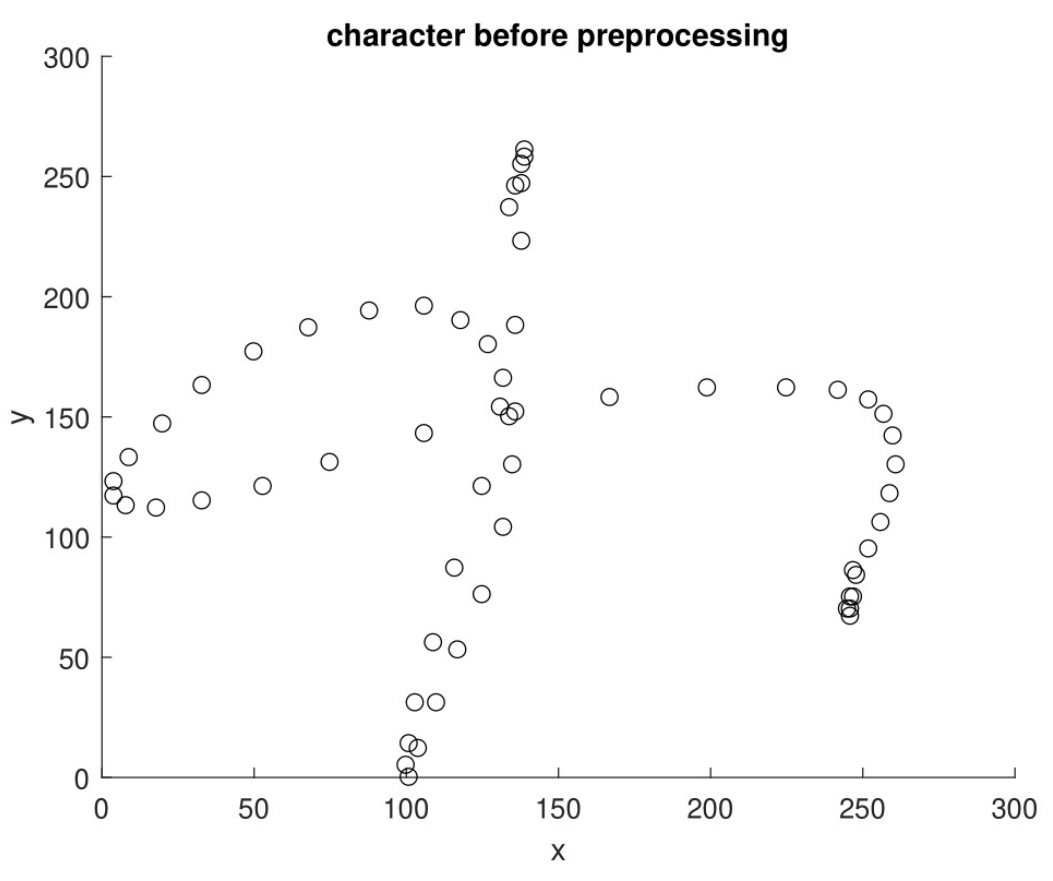}&
\includegraphics[width=2 in]{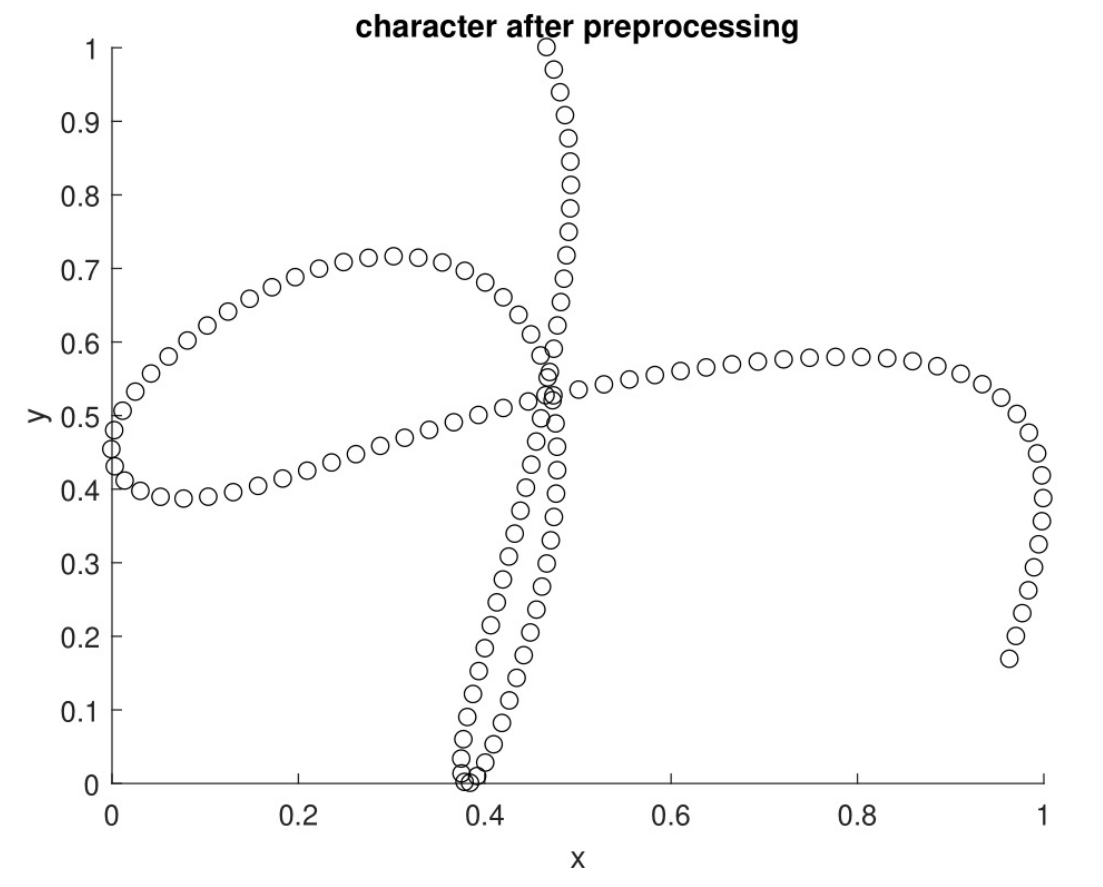}\\
(c)&(d)\\
\end{array}$
\caption{Characters before and after preprocessing. (a) and (c) Characters before preprocessing. (b) and (d) Characters after application of preprocessing steps like removal of repeated points, variations in location and size, variations in distance between consecutive points, and roughness of trace of the characters.}
\end{center}
\label{preproc}
\end{figure}\\
\indent For ST, DFT, DCT, and DWT features, distance $\Delta$ between two consecutive points in a character is chosen so as to have 128 points uniformly spaced along the strokes of the character. The values of distance $\Delta$ between two consecutive points in the characters are 0.0357, 0.0278, and 0.0278 for SP, HOG, and HPOD features, respectively.\\ \indent When the sequences of  x- and y-co-ordinates of points in a character are concatenated, the vector of spatio-temporal features so obtained has dimension of 256. The spans of the character in x- and y-directions are also considered as features. The dimension of the spatio-temporal feature vector after considering the span features becomes 258. These span features are considered in the construction of different feature vectors described in the section on feature extraction. The smallest dimension of the extracted feature vector is 258 and the largest is 786 as explained in the feature extraction section. It is suggested in Duda et al. \cite{troam} to have the number of samples per character class more than the dimension of the feature vectors in that character class. In this work, the average number of samples per character class in the training set is 133 which is smaller than the dimension of the feature vectors to be used for training the classifier. Therefore, the classifier considered for comparison of discriminative capabilities of different features is SVM classifier. SVM classifier can be trained on a small dataset and can produce good classification performance on the corresponding test dataset.             

\section{Feature extraction}
\subsection{Spatio-temporal (ST) features}
\subsubsection{Extraction of ST features from characters}
An online handwritten character is a sequence of strokes, $C=(S_1,\dots,S_i,$\\
$\dots,S_{N_S})$, and a stroke is a sequence of points, as given by (1). Therefore the online  handwritten character is a sequence of points and the corresponding character matrix is represented in terms of these points as \begin{equation}\label{CM}
C^M=[S_1;\dots;S_{N_S}]=[{p_1^{S_1}}^T;\dots;{p_{N_{S_1}}^{S_1}}^T;\dots;{p_n^{S_i}}^T;\dots;{p_{N_C}^{S_{N_S}}}^T],\,\,\, C^M\in[0\,\,\, 1]^{N_C\times 2},\,\,\, N_C=\sum_{i=1}^{N_S}N_{S_i},\end{equation} where $N_C$ is the number of points in the character. $C^M(n,1)$ and $C^M(n,2),\,\,\, 1\le n\le N_C$, are, respectively, the x- and y-co-ordinate values of the $n^{th}$ point in the character. $C^M(*,1)$ and $C^M(*,2)$ are the first and the second columns of the character matrix and are, respectively, the sequences of x- and y-co-ordinates of points in the character. The feature vector in terms of the co-ordinates of points in the character is obtained by concatenating the columns of the character matrix and is obtained as \[X^{ST}=[C^M(*,1)^T, C^M(*,2)^T]^T,\quad X^{ST}\in [0\,\,\,1]^{N_{ST}\times 1},\quad N_{ST}=2\,N_C.\] $X^{ST}$ is the spatio-temporal feature representation of the character.

\subsubsection{Effect of stroke direction and stroke order variations on ST feature representation}
\label{sdod}
\indent Let $C$ be a handwritten character obtained as a sequence of strokes, $C=(S_1,\dots,S_{N_S})$. Let $S_i$ be the $i^{th}$ stroke in $C$ produced by pen movement from the first point towards the last point. First point of the stroke is considered as one of the two end points of the stroke. Let $(x_{-1},y_{-1}),\,\,(x_{o},y_{o})$, and $(x_{1},y_{1})$ be the co-ordinate values of three consecutive points in $C$ as shown in Fig 9(a). Let points $p_{n-1}^{S_i}$, $p_{n}^{S_i}$, and $p_{n+1}^{S_i}$ be three consecutive points in $S_i$ having the co-ordinate values $(x_{-1},y_{-1})$, $(x_{o},y_{o})$, and $(x_{1},y_{1})$, respectively. The locations of $(x_{-1},y_{-1})$, $(x_{o},y_{o})$, and $(x_{1},y_{1})$ in character matrix $C^M$, as given by (2), are, respectively, the rows $r_{n+n_r}^{C,i}=\sum_{i'=1}^{i-1}N_{S_{i'}}+n+n_r,\,\,\, -1\le n_r\le 1$.\\
\indent Let $C'$ be another handwritten character having the same structure and points with the same co-ordinate values as $C$ with all but one stroke produced in the same way as that in $C$. Let the $i^{th}$ stroke $S'_i$ in $C'$ be produced in the opposite direction compared to $S_i$ in $C$. Then, if  $p_{n'}^{S'_i}$ is the point having the co-ordinate value $(x_{o},y_{o})$,  $p_{n'-1}^{S'_i}$ and $p_{n'+1}^{S'_i}$ will have co-ordinate values $(x_{1},y_{1})$ and $(x_{-1},y_{-1})$, respectively. The locations of $(x_{1},y_{1})$, $(x_{o},y_{o})$, and $(x_{-1},y_{-1})$ in the character matrix $C'^M$ are respectively the rows $r_{n'+n_r}^{C',i}=\sum_{i'=1}^{i-1}N_{S'_{i'}}+n'+n_r,\,\,\, -1\le n_r \le 1$. The locations of points $(x_{-1},y_{-1})$, $(x_{o},y_{o})$, and $(x_{1},y_{1})$ in $C^M$ and $C'^M$ are different because the strokes in $C$ and $C'$ in which these points occur have different directions and the same applies to the majority of points in these strokes.\\ 
\indent Let $C''$ be yet another handwritten character having the same structure and points with the same co-ordinate values as $C$. Let $C''$ be produced with different stroke order so that $(i+1)^{th}$ stroke $S''_{i+1}$ in $C''$ is produced as the $i^{th}$ stroke $S_i$ in $C$. Then points $p_{n-1}^{S''_{i+1}}$, $p_{n}^{S''_{i+1}}$, and $p_{n+1}^{S''_{i+1}}$ are points having co-ordinate values $(x_{-1},y_{-1})$, $(x_{o},y_{o})$, and $(x_{1},y_{1})$, respectively. The locations of $(x_{-1},y_{-1})$, $(x_{o},y_{o})$, and $(x_{1},y_{1})$ in $C''^M$ are the rows $r_{n+n_r}^{C'',i+1}=\sum_{i'=1}^{i}N_{S''_{i'}}+n+n_r,\,\,\, -1\le n_r \le 1$, respectively. The locations of the points $(x_{-1},y_{-1})$, $(x_{o},y_{o})$, and $(x_{1},y_{1})$ in $C^M$ and $C''^M$ are different because of the difference in order of the strokes in $C$ and $C''$ in which these points occur and the same applies to all the points in these strokes. Figure 9 illustrates the dependence of character representation and hence the dependence of the location of points in the character matrix on variations in order and direction of strokes in the character.\\
\indent The feature vectors $X^{ST}$, $X'^{ST}$, and $X''^{ST}$ corresponding to the characters $C$, $C'$, and $C''$ respectively are all different because they are constructed from character matrices $C^M$, $C'^M$, and $C''^M$ which are all different from one another although all of them correspond to the characters with the same structure. Therefore ST features are dependent on variations in order and direction of strokes in characters.\\
\begin{center}
\begin{figure}
$\begin{array}{cc}
\includegraphics[height=2.1 in]{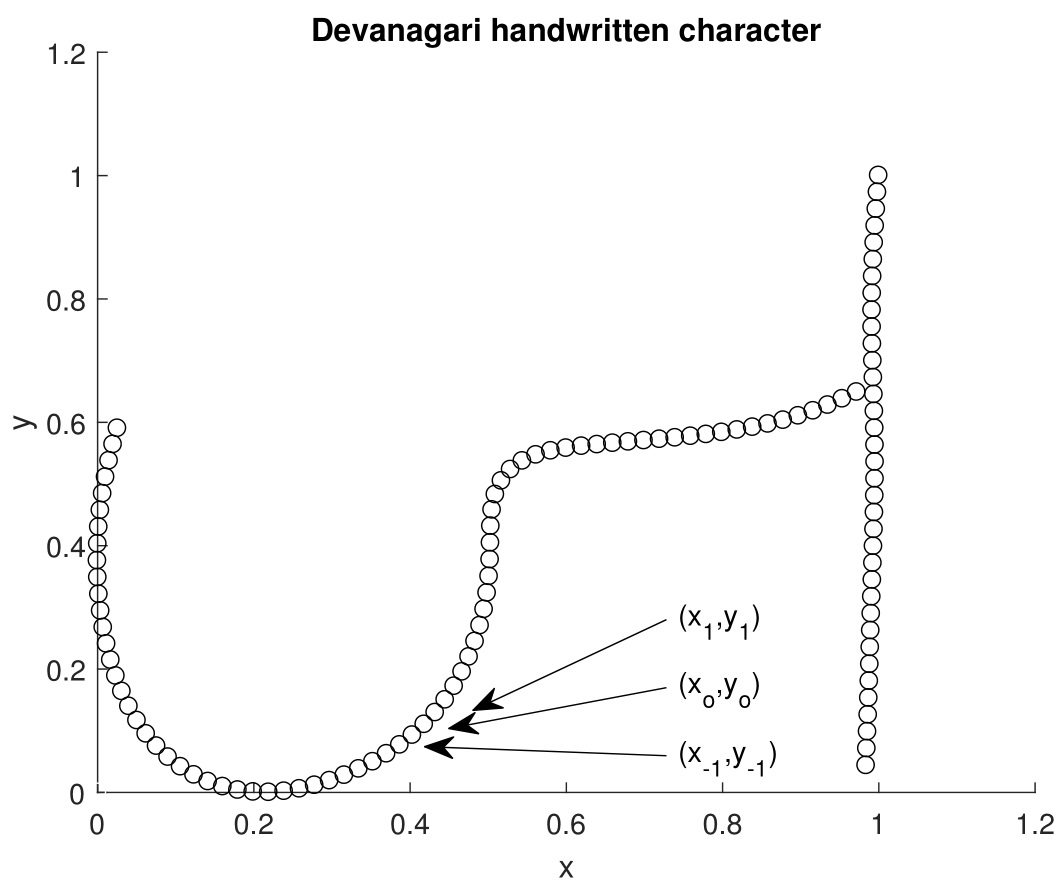}&
\includegraphics[height=2.1 in]{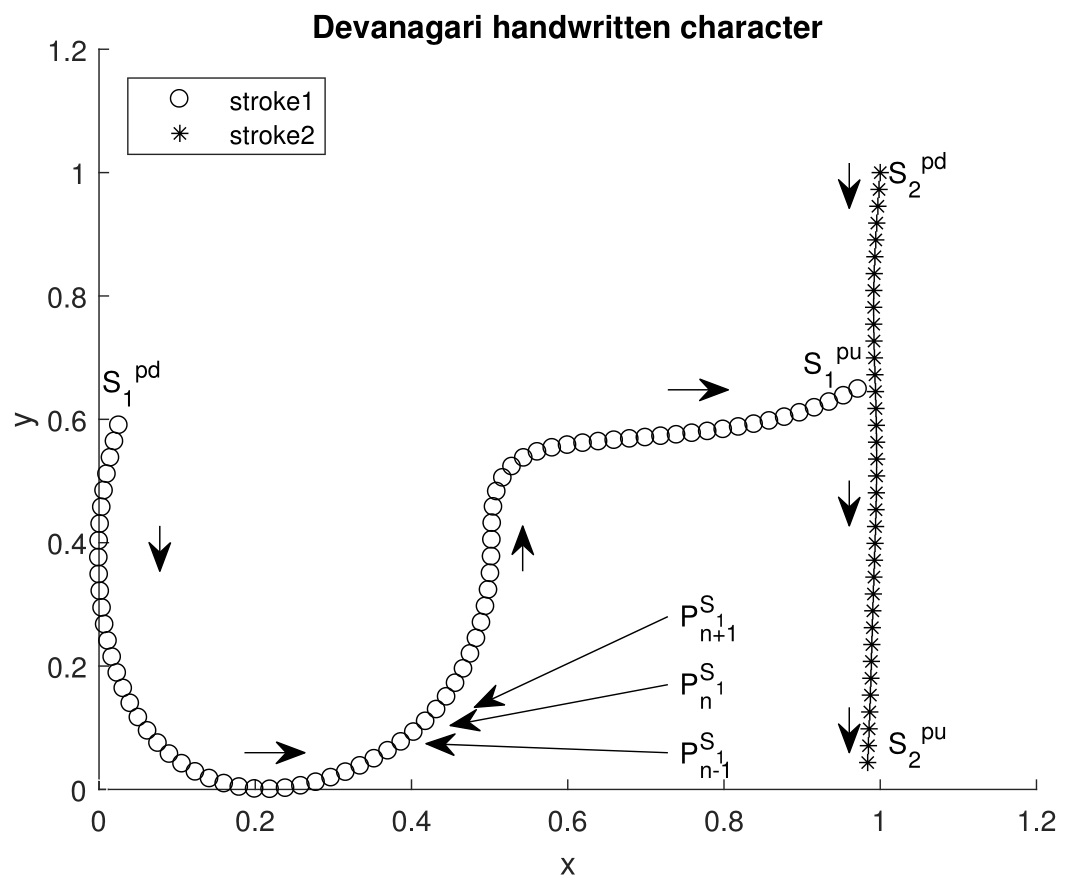}\\
(a)&(b)\\
\includegraphics[height=2.1 in]{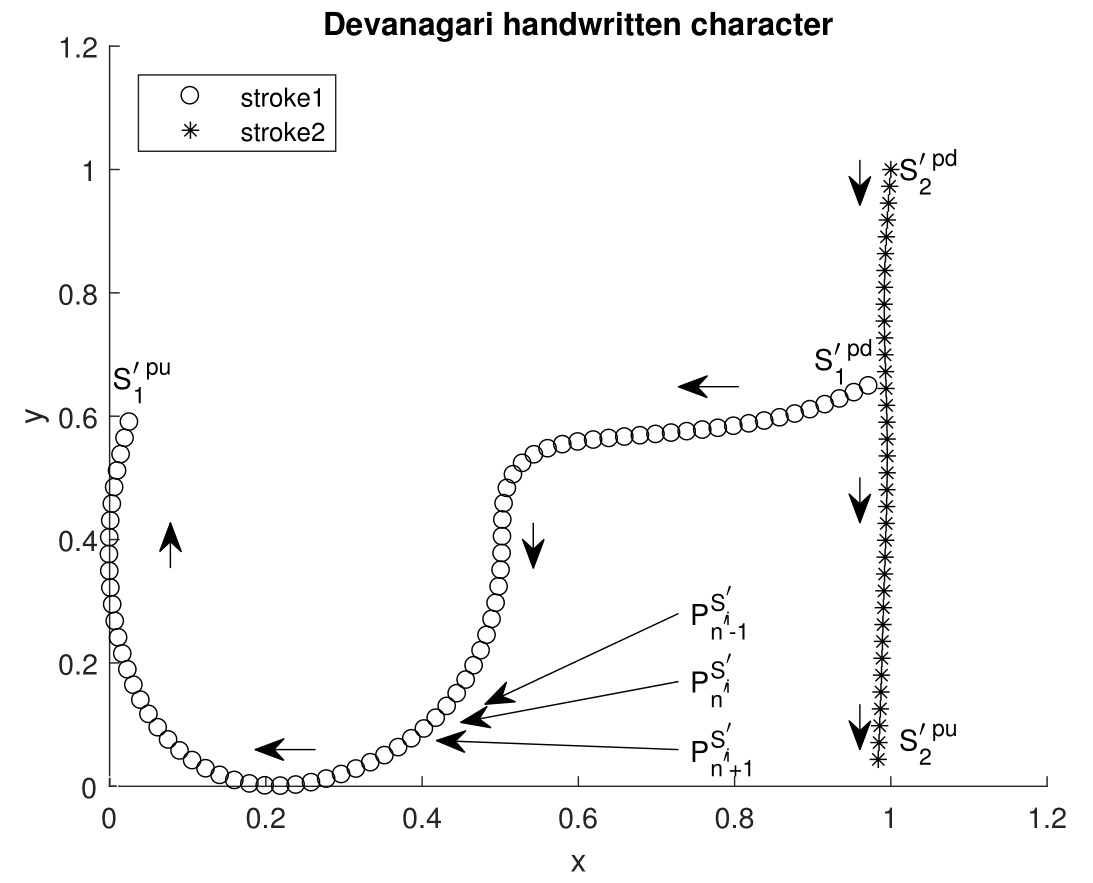}&
\includegraphics[height=2.1 in]{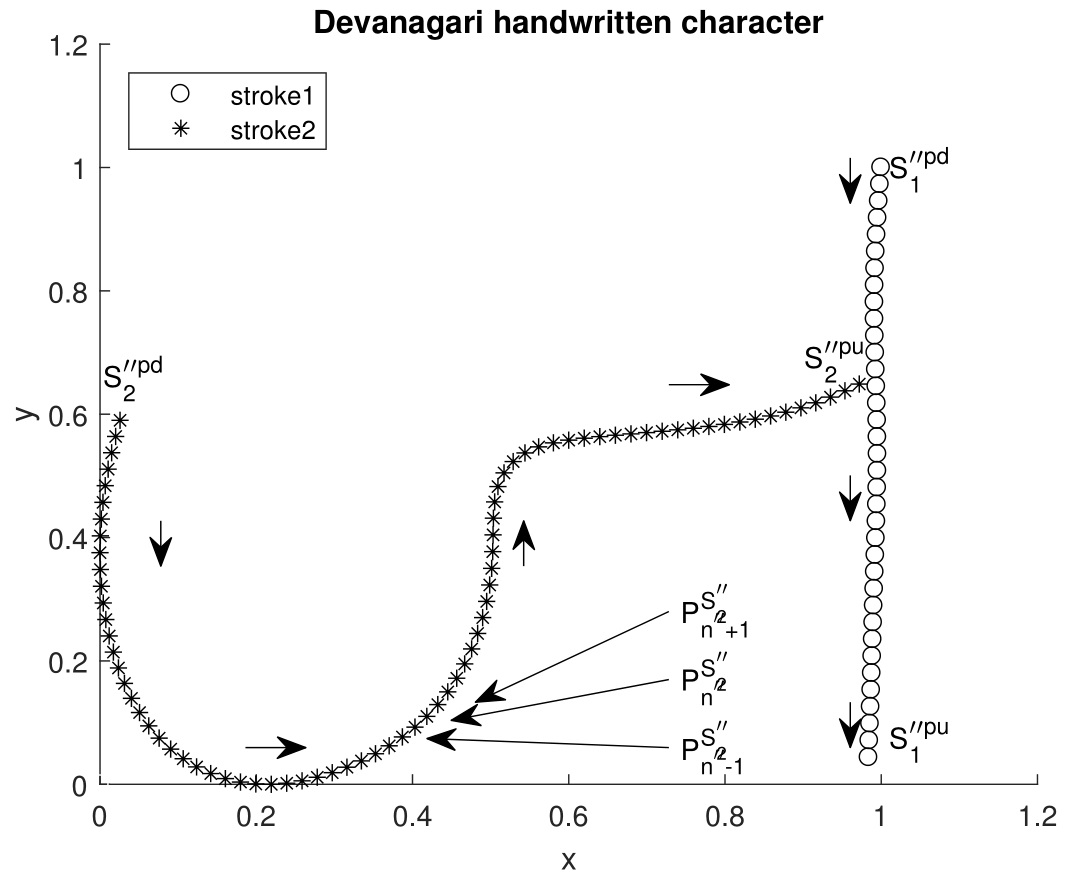}\\
(c)&(d)
\end{array}$
\caption{Illustration of effect of stroke direction and stroke order variations on character representation and hence the character matrix. (a) Hindi online handwritten character. Co-ordinate values of three arbitrary consecutive points in the character are $(x_{-1},y_{-1}),\,\,(x_{o},y_{o}),$ and $(x_{1},y_{1})$. (b) Character in (a) written using two strokes. First stroke (stroke1) is produced by pen movement starting from pen-down at $S_1^{pd}$, moving along the arrows, and ending in pen-up at $S_1^{pu}$. Points $p_{n-1}^{S_1}$, $p_{n}^{S_1},$ and $p_{n+1}^{S_1}$ have co-ordinate values $(x_{-1},y_{-1}),\,\,(x_{o},y_{o}),$ and $(x_{1},y_{1})$, respectively. Second stroke (stroke2) is produced by pen movement starting from pen-down at $S_2^{pd}$, moving down along the arrows, and ending in pen-up at $S_2^{pu}$. (c) Character in (a) written using two strokes. First stroke (stroke1) is produced in opposite direction compared to the first stroke in (b) because of which points corresponding to the co-ordinate values $(x_{1},y_{1}),\,\,(x_{o},y_{o}),$ and $(x_{-1},y_{-1})$ change to $p_{n'-1}^{S'_1}$, $p_{n'}^{S'_1},$ and $p_{n'+1}^{S'_1}$, respectively. Second stroke (stroke2) is produced as in (b). (d) Character in (a) written using two strokes. First stroke (stroke1) is produced as the second stroke in (b). Second stroke (stroke2) is produced as the first stroke in (b) because of which points corresponding to the co-ordinate values $(x_{-1},y_{-1}),\,\,(x_{o},y_{o}),$ and $(x_{1},y_{1})$ change to $p_{{n''}-1}^{{S''}_2}$, $p_{{n''}}^{{S''}_2},$ and $p_{{n''}+1}^{{S''}
_2}$, respectively, $n''=n$.}
\label{sdso}
\end{figure}
\end{center}

\subsection{Discrete Fourier transform (DFT) features}
\subsubsection{Extraction of DFT features from characters}
Discrete Fourier transform (DFT) features are obtained from a character $C$ by constructing a complex vector $C^{cs}$ from the character matrix $C^M$ as $C^{cs}=C^M(*,1)+\jmath\,\,\, C^M(*,2),\,\,\,\sqrt{\jmath}=-1$, $C^{cs}\in\mathfrak{C}^{N_C\times 1}$, $\mathfrak{C}$ is a set of complex numbers. DFT of the vector $C^{cs}$ is obtained as described in Oppenheim et al.  \cite{al},
\[C^{cs}\xrightarrow{DFT}X^{c,DFT},\,\, X^{c,DFT}\in\mathfrak{C}^{N_C\times 1}.\]
Let $X^{r,DFT}=\Re{X^{c,DFT}},\quad X^{i,DFT}=\Im{X^{c,DFT}},\quad X^{r,DFT},X^{i,DFT}\in\mathcal{R}^{N_C\times 1},$ where $\mathcal{R}$ is a set of real numbers. $\Re{X^{c,DFT}}$ and $\Im{X^{c,DFT}}$ are, respectively, the vectors of real and imaginary parts of the vector $X^{c,DFT}$ of complex numbers. The feature vector of DFT features of the character $C$ is obtained as  
\[X^{DFT}=[{X^{r,DFT}}^T,\,{X^{i,DFT}}^T]^T\,,\quad X^{DFT}\in\mathcal{R}^{N_{DFT}\times 1},\quad N_{DFT}=2\,N_C.\]
\subsubsection{Effect of stroke direction and stroke order variations on DFT feature representation}
The DFT features of a character are unique to the character as the character matrix can be recovered from the DFT feature vector. Complex vector $C^{cs}$ can be obtained from the complex vector $X^{c,DFT}$ through the transformation described in Oppenheim et al. \cite{al}\\ 
\[X^{c,DFT}\xrightarrow{IDFT}C^{cs},\,\, C^{cs}\in\mathfrak{C}^{N_C\times 1}.\]IDFT is inverse discrete Fourier transform. Character matrix is constructed from $C^{cs}$ as
\[C^M(*,1)=\Re{C^{cs}},\quad C^M(*,2)=\Im{C^{cs}},\quad C^M\in[0 \,\,\,1]^{N_C\times 2}.\]
$\Re{C^{cs}}$ and $\Im{C^{cs}}$ are, respectively, the vectors of real and imaginary parts of the vector $C^{cs}$ of complex numbers.
If a handwritten character $C'$ having the same structure as $C$ is produced with stroke direction variation or stroke order variation or both then the character matrix $C'^M$ will be different from character matrix $C^M$ as explained in Section 3.1.2. The feature vectors $X^{DFT}$ and $X'^{DFT}$ are constructed from $C^M$ and $C'^M$, respectively, and so $X^{DFT}\ne X'^{DFT}$. So the DFT feature representation is dependent on stroke direction and stroke order variations.
\subsection{Discrete cosine transform (DCT) features}
\subsubsection{Extraction of DCT features from characters}
To extract DCT type II features from a character $C$, vectors $C^{x}$ and $C^{y}$ are, respectively, extracted from the character matrix $C^M$ as
\[C^{x}=C^M(*,1),\quad C^{y}=C^M(*,2),\quad C^x,C^y\in\mathcal{R}^{N_C\times 1}.\] 
DCT of vectors $C^{x}$ and $C^{y}$ are, respectively, obtained as described in Oppenheim et al. \cite{al},
\[C^x\xrightarrow{DCT} X^{x,DCT},\quad C^y\xrightarrow{DCT} X^{y,DCT},\quad X^{x,DCT},X^{y,DCT}\in\mathcal{R}^{N_C\times 1}.\]
The feature vector of DCT features of the character $C$ is obtained as  
\[X^{DCT}=[{X^{x,DCT}}^T,\,\,\,{X^{y,DCT}}^T]^T,\quad X^{DCT}\in\mathcal{R}^{N_{DCT}\times 1},\quad N_{DCT}=2\,N_C.\]

\subsubsection{Effect of stroke direction and stroke order variations on DCT feature representation}
The DCT features of a character are unique to the character as the character matrix can be recovered from the DCT feature vector. The vectors $C^x$ and $C^y$ can be obtained, respectively, from $X^{x,DCT}$ and $X^{y,DCT}$ through the transformation given in Oppenheim et al. \cite{al},
 \[X^{x,DCT}\xrightarrow{IDCT} C^x,\quad X^{y,DCT}\xrightarrow{IDCT} C^y,\quad C^x,C^y\in\mathcal{R}^{N_C\times 1}.\]
IDCT is inverse discrete cosine transform.
Character matrix is constructed from $C^{x}$ and $C^{y}$ as
\[C^M(*,1)=C^{x},\quad C^M(*,2)=C^{y},\quad C^M\in[0 \,\,\,1]^{N_C\times 2}.\]
If a handwritten character $C'$ having the same structure as $C$ is produced with stroke direction variation or stroke order variation or both then the character matrix $C'^M$ will be different from character matrix $C^M$ as explained in Section 3.1.2. The feature vectors $X^{DCT}$ and $X'^{DCT}$ are  constructed from $C^M$ and $C'^M$, respectively, and so  $X^{DCT}\ne X'^{DCT}$ . So the DCT feature representation is dependent on stroke direction and stroke order variations.
\subsection{Discrete wavelet transform (DWT) features}
\subsubsection{Extraction of DWT features from characters}
To extract DWT features from a character $C$, vectors $C^{x}$ and $C^{y}$ are, respectively, obtained from the character matrix $C^M$ as $C^{x}=C^M(*,1),\,\,\, C^{y}=C^M(*,2),\,\,\, C^x,C^y\in\mathcal{R}^{N_C\times 1}.$   DWT of vectors $C^{x}$ and $C^{y}$ are obtained by multi-level decomposition of these vectors using discrete wavelet transform as given in Abbate et al. \cite{ak},
\[C^x\xrightarrow{DWT}X^{xDWT},\quad C^y\xrightarrow{DWT}X^{yDWT},\quad X^{xDWT},X^{yDWT}\in\mathcal{R}^{N_C\times 1}.\]
The feature vector of DWT features of the character $C$ is obtained as 
\[{X}^{DWT}=[{X^{xDWT}}^T,\,\,\,{X^{yDWT}}^T]^T,\quad X^{DWT}\in\mathcal{R}^{N_{DWT}\times 1},\quad N_{DWT}=2N_C.\]

\subsubsection{Effect of stroke direction and stroke order variations on DWT feature representation}
The DWT features of a character are unique to the character as the character matrix can be recovered from the DWT feature vector. vectors $C^x$ and $C^y$ can be obtained from $X^{xDWT}$ and $X^{yDWT}$, respectively, using inverse discrete wavelet transform (IDWT) as described in Abbate et al. \cite{ak},
\[X^{xDWT}\xrightarrow{IDWT}C^x, \quad X^{yDWT}\xrightarrow{IDWT}C^y,\quad C^x,C^y\in\mathcal{R}^{N_C\times 1}.\]
Character matrix is constructed from the vectors $C^x$ and $C^y$ as
\[C^M(*,1)={C^{x}},\quad C^M(*,2)={C^{y}},\quad C^M\in[0 \,\,\,1]^{N_C\times 2}.\]
If a handwritten character $C'$ having the same structure as $C$ is produced with stroke direction variation or stroke order variation or both then the character matrix $C'^M$ will be different from character matrix $C^M$ as explained in Section 3.1.2. The feature vectors $X^{DWT}$ and $X'^{DWT}$ are  constructed from $C^M$ and $C'^M$, respectively, and so  $X^{DWT}\ne X'^{DWT}$. So the DWT feature representation is dependent on stroke direction and stroke order variations.
\subsection{Spatial (SP) features}
\subsubsection{Extraction of SP features from characters}
Spatial features of a character $C$ are obtained by creating a spatial map $C^{sp}$ by quantization of the range of x- and y-co-ordinate values of points in the character as shown in Fig. 10. The spatial quantization step size is $\Delta^{sp}$. The size of $C^{sp}$ is $N_I\times N_I$ where $N_I=({\Delta^{sp}})^{-1}$. The quantization matrix for mapping points in the character to spatial map is obtained as $Q^{sp}(i_1,1)=(i_1-1)\Delta^{sp},\,\,\,Q^{sp}(i_1,2)=i_1\Delta^{sp},\,\,\,1\le i_1\le N_I$. The $n^{th}$ point in the character has x- and y-co-ordinate values $C^M(n,1)$ and $C^M(n,2)$ respectively. The $n^{th}$ point, $1\le\ n\le N_C$, is mapped to $C^{sp}$ by the following considerations:
\[C^{n,i_1}=\begin{cases}1,&Q^{sp}(i_1,1)\le C^M(n,1)<Q^{sp}(i_1,2)\,\,\, \mbox{for some}\,\,\, i_1\in\{1,...,N_I-1\}\\
1,&Q^{sp}(i_1,1)\le C^M(n,1)\le Q^{sp}(i_1,2)\,\,\,\mbox{for}\,\,\, i_1=N_I\\
0,&\mbox{otherwise}\end{cases},\]
\[C^{n,i_2}=\begin{cases}1,&Q^{sp}(i_2,1)\le C^M(n,2)<Q^{sp}(i_2,2)\,\,\, \mbox{for some}\,\,\, i_2\in\{1,...,N_I-1\}\\
1,&Q^{sp}(i_2,1)\le C^M(n,2)\le Q^{sp}(i_2,2)\,\,\,\mbox{for}\,\,\, i_2=N_I\\
0,&\mbox{otherwise}\end{cases}.\]
Then spatial mapping of $n^{th}$ point is done as,
\begin{equation} \label{spmp} C^{sp}(i_1,i_2)=\begin{cases}1,& \,\,\,\mbox{if}\,\,\, I(C^{n,i_1}=1\,\,\, \mbox{and}\,\,\, C^{n,i_2}=1)=1\\
0,&\mbox{otherwise}\end{cases}.\end{equation}
\[I(x)=1\,\,\,\mbox{if}\,\,\, x \,\,\,\mbox{is true and}\,\,\, I(x)=0\,\,\, \mbox{if}\,\,\, x\,\,\,\mbox{is false}.\]  
Let $C^{sp}(*,1),\,\,C^{sp}(*,2),\dots,C^{sp}(*,N_I)$, respectively, be the $1^{st},2^{nd},\dots,N_I^{th}$ columns of $C^{sp}$.  
Then the feature vector of SP features of the character $C$ is obtained by concatenating columns of $C^{sp}$ and is obtained as
\[X^{SP}=[C^{sp}(*,1)^T,\,\,C^{sp}(*,2)^T,\,\,\dots,C^{sp}(*,N_I)^T]^T,\quad X^{SP}\in{\{0,1\}}^{N_{SP}\times 1},\quad N_{SP}=N_I^2.\]
\begin{figure}[ht!]
\begin{center}
$\begin{array}{c}
\includegraphics[height=3 in]{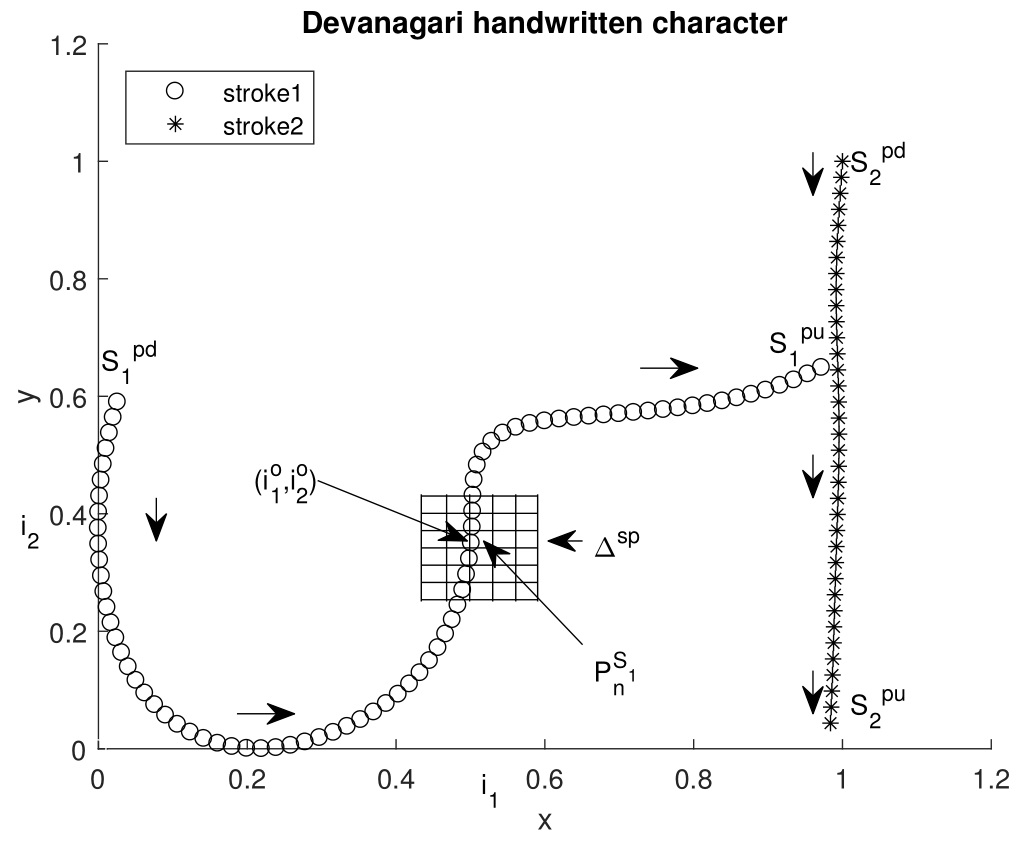}
\end{array}$
\caption{SP features. Range of both the x- and y-co-ordinates of points of online handwritten character $C$ is divided into equal intervals of size $\Delta^{sp}$ to create a spatial map. Intervals of the map in x- and y-directions are indexed by variables $i_1$ and $i_2$, respectively. Part of this map and the corresponding spatial grid is shown, where the x- and y-co-ordinate values of the point $P_n^{S_1}$ are mapped using (3) to the spatial region with indices $i^o_1$ and $i^o_2$, respectively. Similar mapping is applied to all the points in the character to get a spatial representation $C^{sp}$ of the character $C$.}
\label{spmp}
\end{center}
\end{figure}

\subsubsection{Effect of stroke direction and stroke order variations on the SP feature representation}
Let $C$, $C'$, and $C''$ be three handwritten characters and $C^M$, $C'^M$, and $C''^M$ be the corresponding character matrices. Let $(x_{-1},y_{-1})$, $(x_{o},y_{o}),$ and $(x_{1},y_{1})$ be the co-ordinate values of three arbitrary consecutive points in the characters  $C$, $C'$, and $C''$ as described in Section 3.1.2. Let $(i^{-1}_1,i^{-1}_2)$, $(i^{o}_1,i^{o}_2)$, and $(i^{1}_1,i^{1}_2)$, respectively, be the indices of spatial mappings of the aforementioned points. Then spatial mappings of these three consecutive points in the characters $C$, $C'$, and $C''$ are $(C^{sp}(i^{-1}_1,i^{-1}_2),\,C^{sp}(i^{o}_1,i^{o}_2),\,C^{sp}(i^{1}_1,i^{1}_2))$, $(C'^{sp}(i^{-1}_1,i^{-1}_2),\,C'^{sp}(i^{o}_1,i^{o}_2),\,C'^{sp}(i^{1}_1,i^{1}_2)),$ and $({C''}^{sp}(i^{-1}_1,i^{-1}_2),$\\${C''}^{sp}(i^{o}_1,i^{o}_2),\,{C''}^{sp}(i^{1}_1,i^{1}_2))$, respectively, as given by (3).\\ 
\indent The locations of points $(x_{-1},y_{-1})$, $(x_{o},y_{o}),$ and $(x_{1},y_{1})$ in character matrices $C^M$, $C'^M$, and $C''^M$ are different because of the dependence of construction of character matrix on stroke direction and stroke order variations as described in Section 3.1.2. Spatial mapping indices $(i^{-1}_1,i^{-1}_2)$, $(i^{o}_1,i^{o}_2),$ and $(i^{1}_1,i^{1}_2)$ of the points $(x_{-1},y_{-1})$, $(x_{o},y_{o}),$ and $(x_{1},y_{1})$, respectively, in the characters $C$, $C'$, and $C''$, however, are the same. Spatial mapping of points is done using (3) and is dependent on the co-ordinate values of the points rather than on the location of the points in the character matrices. This applies to all the points in all the characters and so the spatial mappings $C^{sp}$, $C'^{sp}$, and $C''^{sp}$ are identical. The SP features are obtained from spatial mappings, so the vector of spatial features $X^{SP}$, $X'^{SP}$, and $X''^{SP}$ of the  characters $C$, $C'$, and $C''$, respectively, are also identical. Therefore SP features are independent of stroke direction and stroke order variations in characters.
\subsection{Histograms of oriented gradients (HOG) features}
\subsubsection{Extraction of HOG features from characters}
HOG features are obtained by determining spatial map $C^{sp}$ of points in a character as described in Section 3.5.1. Gradient of $C^{sp}$ is determined by convolving $C^{sp}$ with gradient filters as described in Dalal et al. \cite{ftrm} to get the magnitude of gradient $C^{g_m}(i_1,i_2)$ and the orientation of gradient $0\le C^{g_{\theta}}(i_1,i_2)\le 180^o,\,\,\, 1\le i_1,i_2\le N_I.$ Orientation of gradient $C^{g_{\theta}}$ is quantized into $N_o$ quantization levels, $N_o=180^o\,{(\Delta^o)}^{-1},$ where $\Delta^o$ is the quantization step size. The corresponding quantization matrix is obtained as $Q^o(q,1)=(q-1)\,\Delta^o, \,\,\, Q^o(q,2)=q\,\Delta^o$, for $1\le q\le N_o$. Histograms of orientation of gradients are obtained by dividing $C^{g_\theta}$ into number of cells $N_{cl}$ of size $\Delta^{cl}=N_I\,N_{cl}^{-1}$ and collecting histogram of  orientation of gradients in each cell to get $C^{h_c(i_3,i_4)},\,\,\,1\le i_3,i_4\le N_{cl}$. Histograms computed in different cells are then collected into overlapping blocks of size $\Delta^b$ with $\Delta^{ov}$ number of overlapping cells between adjacent blocks to get \[C^{h_b(i_5,i_6)},\,\,\, 1\le i_5,i_6\le N_b,\quad N_b=\frac{N_{cl}-\Delta^b}{\Delta^b-\Delta^{ov}}+1,\,\,\, C^{h_b(i_5,i_6)}\in\mathcal{R}^{N_H\times 1},\,\,\,N_H=N_o\,(\Delta^b)^2.\]  
This is shown in Fig. 11 for $\Delta^{ov}=0,\,\,\,\Delta^{b}=1,\,\,\,N_b=N_{cl},\,\,\,i_3=i_5$, and $i_4=i_6$. Then feature vector of HOG features of character $C$ is obtained as
\[X^{HOG}=[{C^{h_b(1,1)}}^T,\dots,{C^{h_b(1,N_b)}}^T,\dots,{C^{h_b(2,1)}}^T,\dots,{C^{h_b(2,N_b)}}^T,\dots,{C^{h_b(N_b,1)}}^T,\dots,{C^{h_b(N_b,N_b)}}^T]^T,\]
$X^{HOG}\in\mathcal{R}^{N_{HOG}\times 1},\,\,\, N_{HOG}=N_H\,N_b^2.$
\begin{center}
\begin{figure}
$\begin{array}{ccc}
\includegraphics[height=2 in]{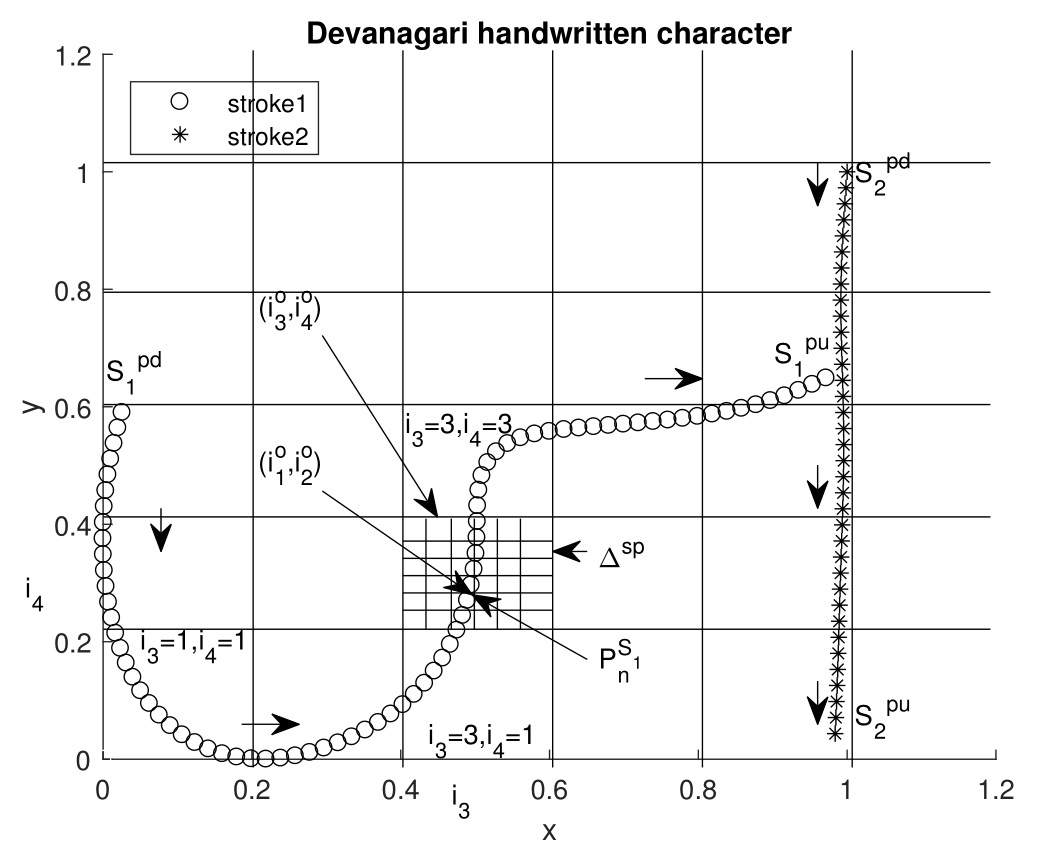}&
\includegraphics[height=2 in]{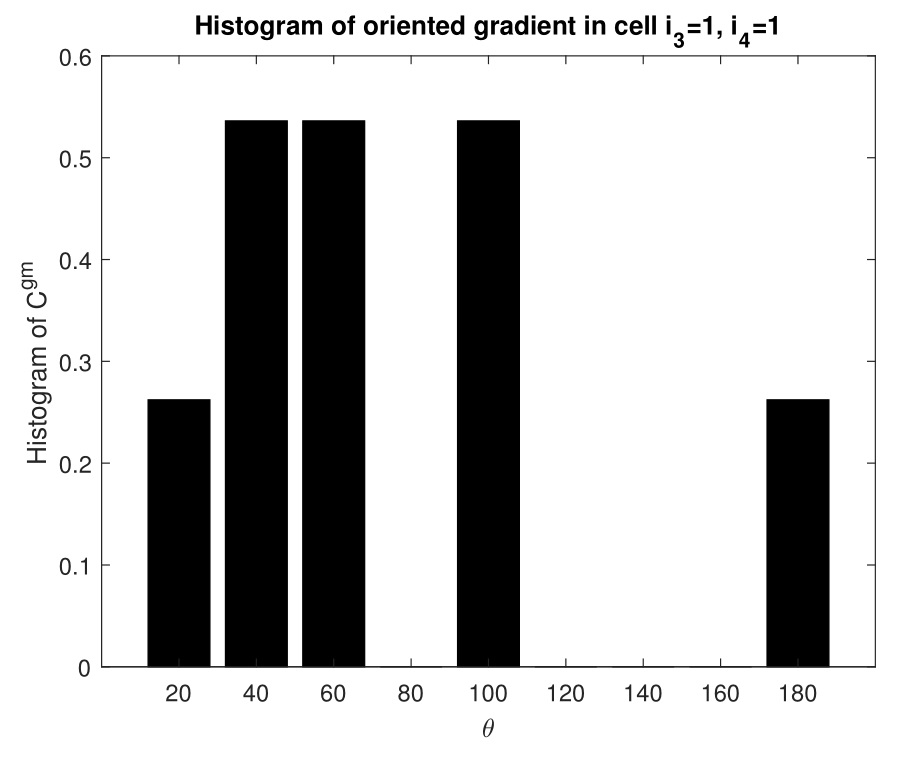}\\
(a)&(b)\\
\includegraphics[height=2 in]{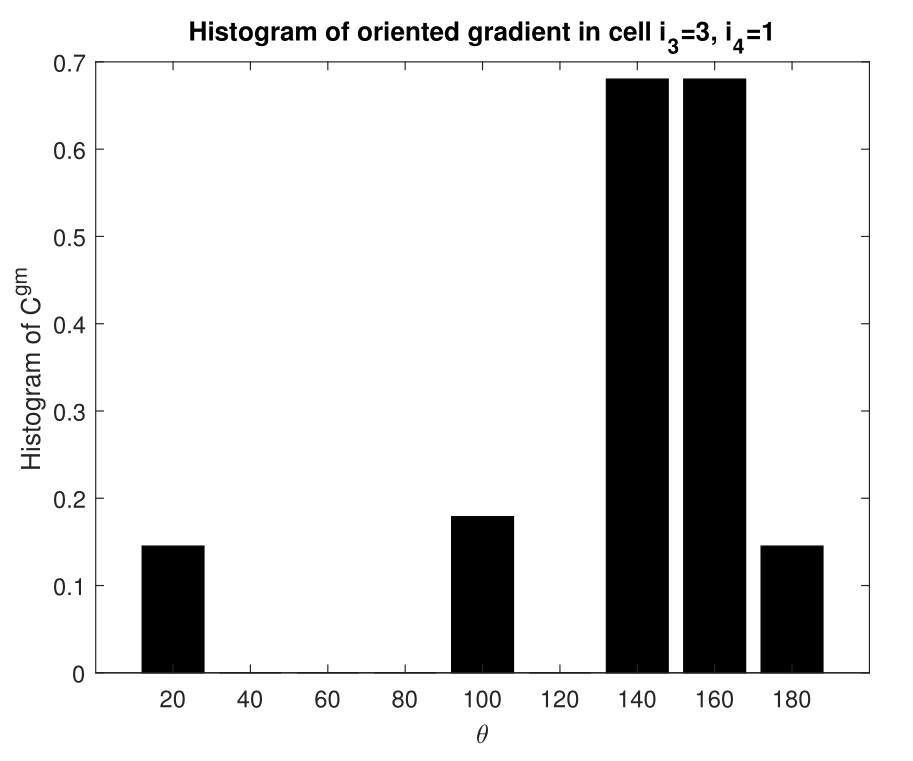}&
\includegraphics[height=2 in]{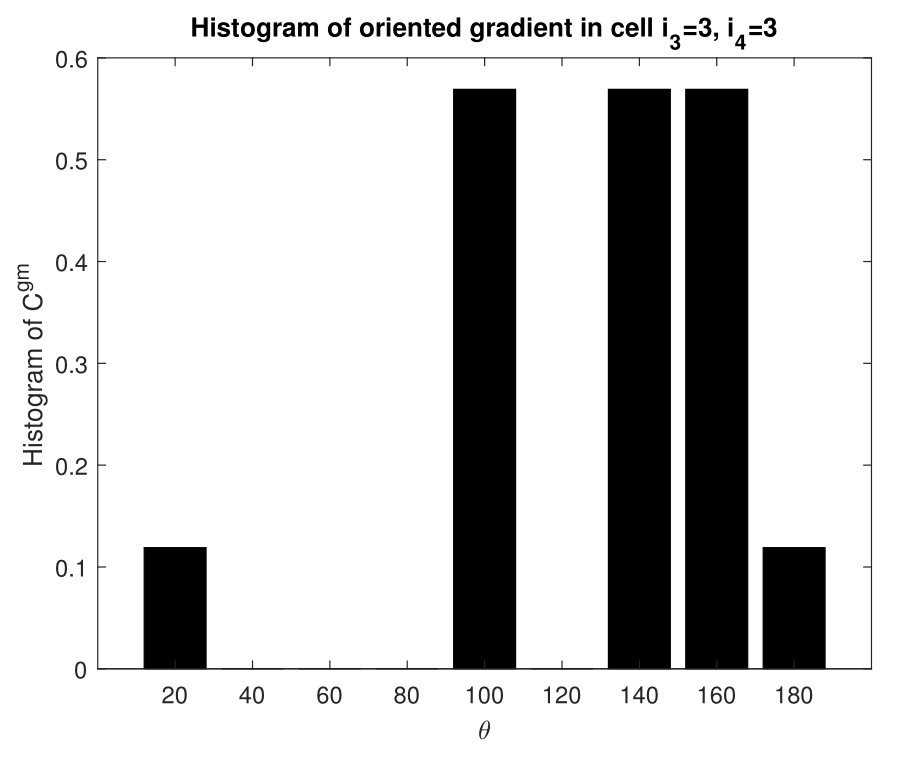}\\
(c)&(d)\\
\end{array}$
\caption{HOG features. (a) Spatial map of the character is divided into cells indexed in x- and y-directions by $i_3$ and $i_4$, respectively. The values of orientation of gradients $\theta$, which are in the range $[0^{\circ}\,\,180^{\circ}]$, are quantized into $9$ intervals each of width $20^{\circ}$. Histogram of orientation of gradients is computed in each of these cells. Cells $(i_3=1,\,\,i_4=1)$, $(i_3=3,\,\,i_4=1)$, and $(i_3=3,\,\,i_4=3)$ are considered for showing HOG features. (b) HOG features in the cell $(i_3=1,\,\,i_4=1)$. (c) HOG features in the cell $(i_3=3,\,\,i_4=1)$. (d) HOG features in the cell $(i_3=3,\,\,i_4=3)$.}
\end{figure}
\end{center}

\subsubsection{Effect of stroke direction and stroke order variations on HOG feature representation}
HOG features are obtained by spatial mapping of points in characters and computation of orientation of gradients followed by collection of histograms of the orientation of gradients from the spatial map. It is observed from Section 3.5.2 that spatial representation of a character is independent of stroke direction and stroke order variations. HOG features are extracted from spatial representation of a character and so are independent of stroke direction and stroke order variations in the character.  
\subsection{Histograms of points, orientations and dynamics of orientations (HPOD) features}
\subsubsection{Extraction of HPOD features from characters}
HPOD features are computed by finding orientations and dynamics of orientations of strokes at points in characters. Character $C$, strokes $S_i,\,\,1\le i\le N_S,$ and points $p_n^{S_i},\,\,1\le n\le N_{S_i},$ are as given in Section 2.4. Direction of stroke $S_i$ at point $p_n^{S_i}$ towards point $p_q^{S_i}$ is \[v_{n,q}^{S_i}=(p_q^{S_i}-p_n^{S_i})(||p_q^{S_i}-p_n^{S_i}||_2)^{-1},\,\,\,v_{n,q}^{S_i}\in\mathcal{R}^{2\times 1},\,\,\,||v_{n,q}^{S_i}||_2=1,\,\,\,1\le n,q\le N_{S_i},\,\,\,1\le i\le N_S.\]
Orientation of stroke $S_i$ at point $p_n^{S_i}$ is \[{\theta'}_n^{O,S_i}=\tan^{-1}(\frac{p_{n+n_{o}}^{S_i}(2)-p_{n-n_{o}}^{S_i}(2)}{p_{n+n_{o}}^{S_i}(1)-p_{n-n_{o}}^{S_i}(1)}),\,\,\,-90^{\circ}\le{\theta'_n}^{O,S_i}\le 90^{\circ}, \,\,\,n_o+1\le n\le N_{S_i}-n_o,\,\,\,1\le i\le N_S,\]
\[\theta_n^{O,S_i}=\begin{cases}{\theta'_n}^{O,S_i},& \mbox{if}\,\,\,p_{n+n_{o}}^{S_i}(1)-p_{n-n_{o}}^{S_i}(1)\ge0\,\,\,\mbox{and}\,\,\,p_{n+n_{o}}^{S_i}(2)-p_{n-n_{o}}^{S_i}(2)\ge0\\
{\theta'_n}^{O,S_i},& \mbox{if}\,\,\,p_{n+n_{o}}^{S_i}(1)-p_{n-n_{o}}^{S_i}(1)<0\,\,\,\mbox{and}\,\,\,p_{n+n_{o}}^{S_i}(2)-p_{n-n_{o}}^{S_i}(2)<0\\
{\theta'_n}^{O,S_i}+180^{\circ},& \mbox{if}\,\,\,p_{n+n_{o}}^{S_i}(1)-p_{n-n_{o}}^{S_i}(1)\ge0\,\,\,\mbox{and}\,\,\,p_{n+n_{o}}^{S_i}(2)-p_{n-n_{o}}^{S_i}(2)<0\\
{\theta'_n}^{O,S_i}+180^{\circ},& \mbox{if}\,\,\,p_{n+n_{o}}^{S_i}(1)-p_{n-n_{o}}^{S_i}(1)<0\,\,\,\mbox{and}\,\,\,p_{n+n_{o}}^{S_i}(2)-p_{n-n_{o}}^{S_i}(2)\ge0\\\end{cases},\]
\begin{equation} 0\le{\theta_n}^{O,S_i}\le 180^{\circ}, \,\,\,n_o+1\le n\le N_{S_i}-n_o,\,\,\,1\le i\le N_S,\,\,\,n_o\ge 1.\end{equation}
 Here $\theta_1^{O,S_i},\dots,\theta_{n_o}^{O,S_i}$ are all made equal to $\theta_{n_o+1}^{O,S_i}$ and $\theta_{N_{S_i}-n_o+1}^{O,S_i},\dots,\theta_{N_{S_i}}^{O,S_i}$ are all made equal to $\theta_{N_{S_i}-n_o}^{O,S_i}$ for $1\le i\le N_S$. Orientations of strokes at points are quantized with quantization step size $\Delta^{\theta_o}$ using quantization matrix $Q^{\theta_o}$. The quantization matrix $Q^{\theta_o}$ is constructed as 
\[Q^{\theta_o}(q_1,1)=(q_1-1)\Delta^{\theta_o},\,\,\,Q^{\theta_o}(q_1,2)=q_1\Delta^{\theta_o},\,\,\,1\le q_1\le N_{\theta_o},\,\,\,N_{\theta_o}=180^{\circ}(\Delta^{\theta_o})^{-1}.\]
Change of direction of stroke $S_i$ at point $p_n^{S_i}$ or dynamics of orientation of stroke $S_i$ at point $p_n^{S_i}$ is 
\begin{equation}\theta_n^{D,S_i}=\cos^{-1}({v_{n-n_{d},n}^{S_i}}^T v_{n,n+n_{d}}^{S_i}),\,\,\,0\le \theta_n^{D,S_i} \le 180^{\circ},\,\,\,n_d+1\le n\le N_{S_i}-n_d,\,\,n_d\ge 1,\,\,1\le i\le N_S.\end{equation}    
Here $\theta_1^{D,S_i},\dots,\theta_{n_d}^{D,S_i}$ are all made equal to $\theta_{n_d+1}^{D,S_i}$ and $\theta_{N_{S_i}-n_d+1}^{D,S_i},\dots,\theta_{N_{S_i}}^{D,S_i}$ are all made equal to $\theta_{N_{S_i}-n_d}^{D,S_i}$ for $1\le i\le N_S$. The values of dynamics of orientations of stroke at points are quantized with quantization step size $\Delta^{\theta_d}$  using quantization matrix $Q^{\theta_d}$. The quantization matrix  $Q^{\theta_d}$ is constructed as
\[Q^{\theta_d}(q_2,1)=(q_2-1)\Delta^{\theta_d},\,\,\,Q^{\theta_d}(q_2,2)=q_2\Delta^{\theta_d},\,\,\,1\le q_2\le N_{\theta_d},\,\,\,N_{\theta_d}=180^{\circ}(\Delta^{\theta_d})^{-1}.\]
Orthogonal orientation to stroke $S_i$ at point $p_n^{S_i}$ is 
\[{\theta'}_n^{\perp,S_i}=-\tan^{-1}(\frac{p_{n+n_{o}}^{S_i}(1)-p_{n-n_{o}}^{S_i}(1)}{p_{n+n_{o}}^{S_i}(2)-p_{n-n_{o}}^{S_i}(2)}),\,\,\,-90^{\circ}\le{\theta'_n}^{O,S_i}\le 90^{\circ}, \,\,\,n_o+1\le n\le N_{S_i}-n_o,\,\,\,1\le i\le N_S.\]
${\theta}_n^{\perp,S_i}$ is obtained from ${\theta'}_n^{\perp,S_i}$ the same way as ${\theta}_n^{O,S_i}$ is obtained from ${\theta'}_n^{O,S_i}$ so that $0\le\theta_n^{\perp,S_i}\le 180^{\circ}, \,\,\,n_o+1\le n\le N_{S_i}-n_o,\,\,\,1\le i\le N_S.$ Here $\theta_1^{\perp,S_i},\dots,\theta_{n_o}^{\perp,S_i}$ are all made equal to $\theta_{n_o+1}^{\perp,S_i} \,\,\,\mbox{and}$\\ $\theta_{N_{S_i}-n_o+1}^{\perp,S_i},\dots,\theta_{N_{S_i}}^{\perp,S_i}$ are all made equal to $\theta_{N_{S_i}-n_o}^{\perp,S_i}$ for $1\le i\le N_S.$\\
\indent The co-ordinate values, the orientations and the dynamics of orientations of strokes $S_i$ at points $p_n^{S_i},\,\,\,1\le n\le N_{S_i},\,\,\,1\le i\le N_S$, are spatially mapped which are represented by matrices $C^{p,sp}$, $C^{o,sp}$, and $C^{d,sp}$ respectively. The mapping is done using spatial quantization matrix $Q^{sp}$ constructed as $Q^{sp}(i_1,1)=(i_1-1)\Delta^{sp},\,\,\,Q^{sp}(i_1,2)=i_1\Delta^{sp},\,\,\,1\le i_1\le N_I,\,\,\, N_I=(\Delta^{sp})^{-1},\,\,\,\Delta^{sp}\approx\Delta$. The spatial quantization step size is made approximately equal to the distance between consecutive points in the character so as to have approximately one point in the character mapped to a particular spatial grid region. Spatial mapping of co-ordinate values of points $p_n^{S_i}$ of strokes $S_i$ of character $C$, $1\le n\le N_{S_i},\,\,\,1\le i\le N_S$, is done by the following considerations:
\[C^{n,i_1}=\begin{cases}1,&Q^{sp}(i_1,1)\le p_n^{S_i}(1,1)<Q^{sp}(i_1,2)\,\,\, \mbox{for some}\,\,\, i_1\in\{1,...,N_I-1\}\\
1,&Q^{sp}(i_1,1)\le p_n^{S_i}(1,1)\le Q^{sp}(i_1,2)\,\,\,\mbox{for}\,\,\, i_1=N_I\\
0,&\mbox{otherwise}\end{cases},\]
\[C^{n,i_2}=\begin{cases}1,&Q^{sp}(i_2,1)\le p_n^{S_i}(2,1)<Q^{sp}(i_2,2)\,\,\, \mbox{for some}\,\,\, i_2\in\{1,...,N_I-1\}\\
1,&Q^{sp}(i_2,1)\le p_n^{S_i}(2,1)\le Q^{sp}(i_2,2)\,\,\,\mbox{for}\,\,\, i_2=N_I\\
0,&\mbox{otherwise}\end{cases}. \]
Then spatial mapping of co-ordinate values of $p_n^{S_i}$ is done as,
\[C^{p,sp}(i_1,i_2)=\begin{cases}1,& \,\,\,\mbox{if}\,\,\, I(C^{n,i_1}=1\,\,\, \mbox{and}\,\,\, C^{n,i_2}=1)=1\\
0,&\mbox{otherwise}\end{cases}.\]  
\indent Stroke thickening increases the character discriminative capability of HPOD features. Stroke thickening at a point $p_n^{S_i}$ is done spatially by introducing points on either side of the point in the direction orthogonal to the orientation of the stroke at that point. Spatial mapping $C^{p,sp}(i_1,i_2)$ of point $p_n^{S_i}$ has at most eight nearest grid regions available where points can be introduced for spatial stroke thickening. The values of $\theta_n^{\perp,S_i}$ are therefore quantized to realize spatial thickening of stroke at $p_n^{S_i}$ in the following way:
\[C^{n,i_1,i_2,1}=\begin{cases}1,\quad i_1^{1,1}=i_1-1,i_1^{1,2}=i_1+1,i_2^{1,1}=i_2,i_2^{1,2}=i_2,&\mbox{if}\,\,\,0^o\le\theta_n^{\perp,S_i}<22.5^o\\
1,\quad i_1^{1,1}=i_1-1,i_1^{1,2}=i_1+1,i_2^{1,1}=i_2,i_2^{1,2}=i_2,&\mbox{if}\,\,\,157.5^o\le \theta_n^{\perp,S_i}\le 180^o\\0,&\mbox{otherwise}\end{cases},\]
\[C^{n,i_1,i_2,2}=\begin{cases}1,\quad i_1^{2,1}=i_1-1,i_1^{2,2}=i_1+1,i_2^{2,1}=i_2-1,i_2^{2,2}=i_2+1,&\mbox{if}\,\,\,22.5^o\le\theta_n^{\perp,S_i}<67.5^o\\
0,&\mbox{otherwise}\end{cases},\]
\[C^{n,i_1,i_2,3}=\begin{cases}1,\quad i_1^{3,1}=i_1,i_1^{3,2}=i_1,i_2^{3,1}=i_2-1,i_2^{3,2}=i_2+1,&\mbox{if}\,\,\,67.5^o\le\theta_n^{\perp,S_i}<112.5^o\\
0,&\mbox{otherwise}\end{cases},\]
\[C^{n,i_1,i_2,4}=\begin{cases}1,\quad i_1^{4,1}=i_1-1,i_1^{4,2}=i_1+1,i_2^{4,1}=i_2+1,i_2^{4,2}=i_2-1,&\mbox{if}\,\,\,112.5^o\le\theta_n^{\perp,S_i}<157.5^o\\
0,&\mbox{otherwise}\end{cases}.\]
Stroke thickening at $p_n^{S_i},\,\,\,1\le n\le N_{S_i},\,\,\,1\le i\le N_S$, is done spatially as,
\[C^{p,sp}(i_1^{l,1},i_2^{l,1})=1,\,\,\,C^{p,sp}(i_1^{l,2},i_2^{l,2})=1\,\,\,\mbox{if}\,\,\,C^{n,i_1,i_2,l}=1\,\,\,\mbox{for some}\,\,\,l\in\{1,2,3,4\}.\]   
Spatial mapping of orientation of stroke at $p_n^{S_i},\,\,\,1\le n\le N_{S_i},\,\,\,1\le i\le N_S,$ is done as,\[
C^{o,sp}(i_1,i_2)=\begin{cases}\theta_n^{O,S_i},& \,\,\,\mbox{if}\,\,\, I(C^{n,i_1}=1\,\,\, \mbox{and}\,\,\, C^{n,i_2}=1)=1\\
0,&\mbox{otherwise}\end{cases}.\]
Thickening of orientations of strokes at $p_n^{S_i},\,\,\,1\le n\le N_{S_i},\,\,\,1\le i\le N_S,$ is done spatially as,
\[C^{o,sp}(i_1^{l,1},i_2^{l,1})=\theta_n^{O,S_i},\,\,\,C^{o,sp}(i_1^{l,2},i_2^{l,2})=\theta_n^{O,S_i}\,\,\,\mbox{if}\,\,\,C^{n,i_1,i_2,l}=1\,\,\,\mbox{for some}\,\,\,l\in\{1,2,3,4\}.\]   
Spatial mapping of dynamics of orientations of strokes at $p_n^{S_i},\,\,\,1\le n\le N_{S_i},\,\,\,1\le i\le N_S,$ is done as,
\[C^{d,sp}(i_1,i_2)=\begin{cases}\theta_n^{D,S_i},& \,\,\,\mbox{if}\,\,\, I(C^{n,i_1}=1\,\,\, \mbox{and}\,\,\, C^{n,i_2}=1)=1\\
0,&\mbox{otherwise}\end{cases}.\]
Thickening of dynamics of orientations of strokes at $p_n^{S_i},\,\,\,1\le n\le N_{S_i},\,\,\,1\le i\le N_S,$ is done spatially as,
\[C^{d,sp}(i_1^{l,1},i_2^{l,1})=\theta_n^{D,S_i},\,\,\,C^{d,sp}(i_1^{l,2},i_2^{l,2})=\theta_n^{D,S_i}\,\,\,\mbox{if}\,\,\,C^{n,i_1,i_2,l}=1\,\,\,\mbox{for some}\,\,\,l\in\{1,2,3,4\}.\]   
Histograms of points of strokes of character is obtained by dividing $C^{p,sp}$ into number of overlapping cells $N_{pcl}$ of cell size $\Delta^{pcl}=N_I\,N_{pcl}^{-1}$ overlapping by $\Delta_p^{ov}$ . Let $C^{h,p(i_3,i_4)}\in 0^{2\times 1},\,\,\,1\le i_3,i_4\le N_{pcl}$, be array of vectors of zeros, one vector for each cell for collecting distribution of points in each cell. The Histograms of points are obtained by the following considerations:
\[C^{p,i_3,i_1}=\begin{cases}1,& \mbox{for}\,\,\,(i_3-1)\Delta^{pcl}-\Delta_p^{ov}I(i_3>1)+1\le i_1\le i_3\,\Delta^{pcl}+\Delta_p^{ov}I(i_3<N_{pcl})\\
0,& \mbox{otherwise}\end{cases},\]
\[C^{p,i_4,i_2}=\begin{cases}1,& \mbox{for}\,\,\,(i_4-1)\Delta^{pcl}-\Delta_p^{ov}I(i_4>1)+1\le i_2\le i_4\,\Delta^{pcl}+\Delta_p^{ov}I(i_4<N_{pcl})\\
0,& \mbox{otherwise}\end{cases},\]
\[C^{p,i_1,i_2}_1=\begin{cases}1,&\mbox{if}\,\,\, C^{p,sp}(i_1,i_2)=1\\ 
0,& \mbox{otherwise}\end{cases},\]
\[C^{p,i_1,i_2}_2=\begin{cases}1,&\mbox{if}\,\,\, C^{p,sp}(i_1,i_2)=0\\ 
0,& \mbox{otherwise}\end{cases}, \]
\[\mbox{for}\,\,\, 1\le i_1,i_2\le N_I,\quad1\le i_3,i_4\le N_{pcl}.\] 
Histogram of points for cell $C^{p,sp}(i_3,i_4),\,\,\,1\le i_3,i_4\le N_{pcl}$, is (Fig. 12(d)-(f))
\[C'^{h,p(i_3,i_4)}_q=\sum_{i_1}\sum_{i_2}\, I(C^{p,i_3,i_1}=1\,\,\, \mbox{and}\,\,\, C^{p,i_4,i_2}=1 \,\,\,\mbox{and}\,\,\, C^{p,i_1,i_2}_q=1),\] 
\[\mbox{for}\,\,\, 1\le q\le 2.\,\,\,C^{h,p(i_3,i_4)}_q=C'^{h,p(i_3,i_4)}_q\,(\Delta^{pcl})^{-2},\,\,\, C^{h,p(i_3,i_4)}\in\mathcal{R}^{2\times 1}.\]
Vector of histograms of points of character is obtained as
\[C^{h,p}=[{C^{h,p(1,1)}}^T,\dots,{C^{h,p(N_{pcl},1)}}^T,{C^{h,p(1,2)}}^T,\dots,{C^{h,p(N_{pcl},2)}}^T,\dots,{C^{h,p(N_{pcl},N_{pcl})}}^T]^T,\] 
\[C^{h,p}\in\mathcal{R}^{N_{hp}\times1},\,\,\,N_{hp}=2N_{pcl}^2.\]
Histograms of orientations of strokes at points of a character are obtained by dividing $C^{o,sp}$ into number of overlapping cells $N_{ocl}$ of cell size $\Delta^{ocl}=N_I\,N_{ocl}^{-1}$ overlapping by $\Delta_o^{ov}$. Let $C^{h,o(i_3,i_4)}\in 0^{N_{\theta_o}\times 1},\,\,\,1\le i_3,i_4\le N_{ocl}$, be array of vectors of zeros, one vector for each cell for collecting histogram of orientations in each cell. The Histograms of orientations of strokes at points are obtained by the following considerations:
\[C^{o,i_3,i_1}=\begin{cases}1,& \mbox{for}(i_3-1)\Delta^{ocl}-I(i_3>1)\Delta_o^{ov}+1\le i_1\le i_3\,\Delta^{ocl}+I(i_3<N_{ocl})\Delta_o^{ov}\\
0,& \mbox{otherwise}\end{cases},\]
\[C^{o,i_4,i_2}=\begin{cases}1,& \mbox{for}(i_4-1)\Delta^{ocl}-I(i_4>1)\Delta_o^{ov}+1\le i_2\le i_4\,\Delta^{ocl}+I(i_4<N_{ocl})\Delta_o^{ov}\\
0,& \mbox{otherwise}\end{cases},\]
\[C^{o,i_1,i_2}_{q_1}=\begin{cases}1,& Q^{\theta_o}(q_1,1)\le C^{o,sp}(i_1,i_2)<Q^{\theta_o}(q_1,2)\,\,\, \mbox{for some}\,\,\,q_1\in\{1,\dots,N_{\theta_o}-1\}\\
1,& Q^{\theta_o}(q_1,1)\le C^{o,sp}(i_1,i_2)\le Q^{\theta_o}(q_1,2)\,\,\, \mbox{for}\,\,\,q_1=N_{\theta_o}\\
0,& \mbox{otherwise}\end{cases},\]
\[\mbox{for}\,\,\, 1\le i_1,i_2\le N_I,\quad1\le i_3,i_4\le N_{ocl},\quad1\le q_1\le N_{\theta_o}.\]
Histogram of orientations for cell $C^{o,sp}(i_3,i_4),\,\,\,1\le i_3,i_4\le N_{ocl}$, is (Fig. 12(g)-(i))
\[C'^{h,o(i_3,i_4)}_{q_1}=\sum_{i_1}\sum_{i_2}\, I(C^{o,i_3,i_1}=1\,\,\, \mbox{and}\,\,\, C^{o,i_4,i_2}=1 \,\,\,\mbox{and}\,\,\, C^{o,i_1,i_2}_{q_1}=1),\]
\[1\le q_1\le N_{\theta_o}.\,\,\,C^{h,o(i_3,i_4)}=C'^{h,o(i_3,i_4)}\,(||C'^{h,o(i_3,i_4)}||_2+\epsilon_o)^{-1},\,\,\,\epsilon_o>0,\,\,\,C^{h,o(i_3,i_4)}\in\mathcal{R}^{N_{\theta_o}\times 1}.\]
Vector of histograms of orientations of a character is obtained as
\[C^{h,o}=[{C^{h,o(1,1)}}^T,\dots,{C^{h,o(N_{ocl},1)}}^T,{C^{h,o(1,2)}}^T,\dots,{C^{h,o(N_{ocl},2)}}^T,\dots,{C^{h,o(N_{ocl},N_{ocl})}}^T]^T,\] 
\[C^{h,o}\in\mathcal{R}^{N_{ho}\times1},\,\,\,N_{ho}=N_{\theta_o}\,N_{ocl}^2.\]
Histograms of dynamics of orientations of strokes at points of a character are obtained by dividing $C^{d,sp}$ into number of overlapping cells $N_{dcl}$ of cell size $\Delta^{dcl}=N_I\,N_{dcl}^{-1}$ overlapping by $\Delta_d^{ov}$. Let $C^{h,d(i_3,i_4)}\in 0^{N_{\theta_d}\times 1},\,\,\,1\le i_3,i_4\le N_{dcl}$ be array of vectors of zeros, one vector for each cell for collecting histogram of dynamics of orientations in each cell. The Histograms of dynamics of orientations of strokes at points are obtained by the following considerations:
\[C^{d,i_3,i_1}=\begin{cases}1,& \mbox{for}(i_3-1)\Delta^{dcl}-I(i_3>1)\Delta_d^{ov}+1\le i_1\le i_3\,\Delta^{dcl}+I(i_3<N_{dcl})\Delta_d^{ov}\\
0,& \mbox{otherwise}\end{cases},\]
\[C^{d,i_4,i_2}=\begin{cases}1,& \mbox{for}(i_4-1)\Delta^{dcl}-I(i_4>1)\Delta_d^{ov}+1\le i_2\le i_4\,\Delta^{dcl}+I(i_4<N_{dcl})\Delta_d^{ov}\\
0,& \mbox{otherwise}\end{cases},\]
\[C^{d,i_1,i_2}_{q_2}=\begin{cases}1,& Q^{\theta_d}(q_2,1)\le C^{d,sp}(i_1,i_2)<Q^{\theta_d}(q_2,2)\,\,\, \mbox{for some}\,\,\,q_2\in\{1,\dots,N_{\theta_d}-1\}\\
1,& Q^{\theta_d}(q_2,1)\le C^{d,sp}(i_1,i_2)\le Q^{\theta_d}(q_2,2)\,\,\, \mbox{for}\,\,\,q_2=N_{\theta_d}\\
0,& \mbox{otherwise}\end{cases},\]
\[\mbox{for}\,\,\, 1\le i_1,i_2\le N_I,\quad1\le i_3,i_4\le N_{dcl},\quad1\le q_2\le N_{\theta_d}.\]
Histogram of dynamics of orientations for cell $C^{d,sp}(i_3,i_4),\,\,\,1\le i_3,i_4\le N_{dcl}$, is (Fig. 12(j)-(l))
\[C'^{h,d(i_3,i_4)}_{q_2}=\sum_{i_1}\sum_{i_2}\, I(C^{d,i_3,i_1}=1\,\,\, \mbox{and}\,\,\, C^{d,i_4,i_2}=1 \,\,\,\mbox{and}\,\,\, C^{d,i_1,i_2}_{q_2}=1),\]
\[\mbox{for}\,\,\, 1\le q_2\le N_{\theta_d}.\,\,\,C^{h,d(i_3,i_4)}=C'^{h,d(i_3,i_4)}\,(||C'^{h,d(i_3,i_4)}||_2+\epsilon_d)^{-1},\,\,\,\epsilon_d>0,\,\,\,C^{h,d(i_3,i_4)}\in\mathcal{R}^{N_{\theta_d}\times 1}.\]
Vector of histograms of dynamics of orientations of character is obtained as
\[C^{h,d}=[{C^{h,d(1,1)}}^T,\dots,{C^{h,d(N_{dcl},1)}}^T,{C^{h,d(1,2)}}^T,\dots,{C^{h,d(N_{dcl},2)}}^T,\dots,{C^{h,d(N_{dcl},N_{dcl})}}^T]^T,\] 
\[C^{h,d}\in\mathcal{R}^{N_{hd}\times1},\,\,\,N_{hd}=N_{\theta_d}\,N_{dcl}^2.\]
Then the feature vector of HPOD features of character $C$ is obtained as
\[X^{HPOD}=[{C^{h,p}}^T,{C^{h,o}}^T,{C^{h,d}}^T]^T,\,\,\,X^{HPOD}\in\mathcal{R}^{N_{HPOD}\times 1},\,\,\,N_{HPOD}=N_{hp}+N_{ho}+N_{hd}.\]
\begin{center}
\begin{figure}
$\begin{array}{ccc}
\includegraphics[height=1.2 in]{ohwrjaspp2.jpg}&
\includegraphics[height=1.2 in]{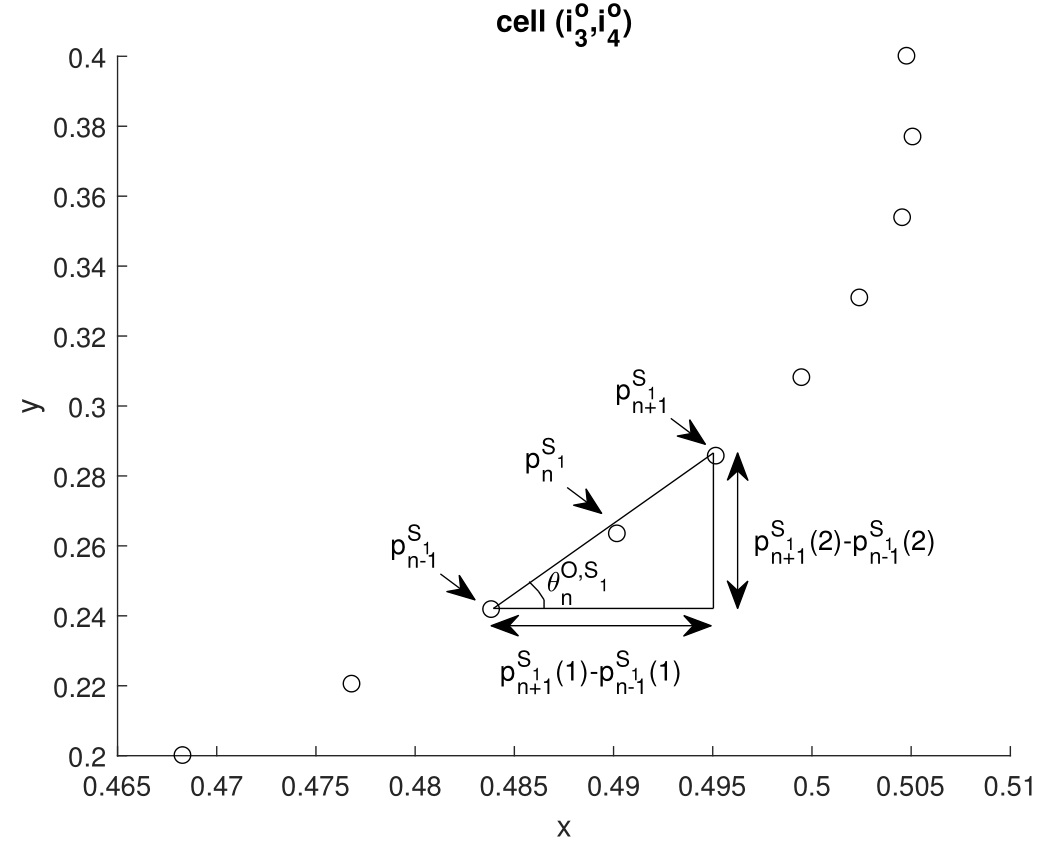}&
\includegraphics[height=1.2 in]{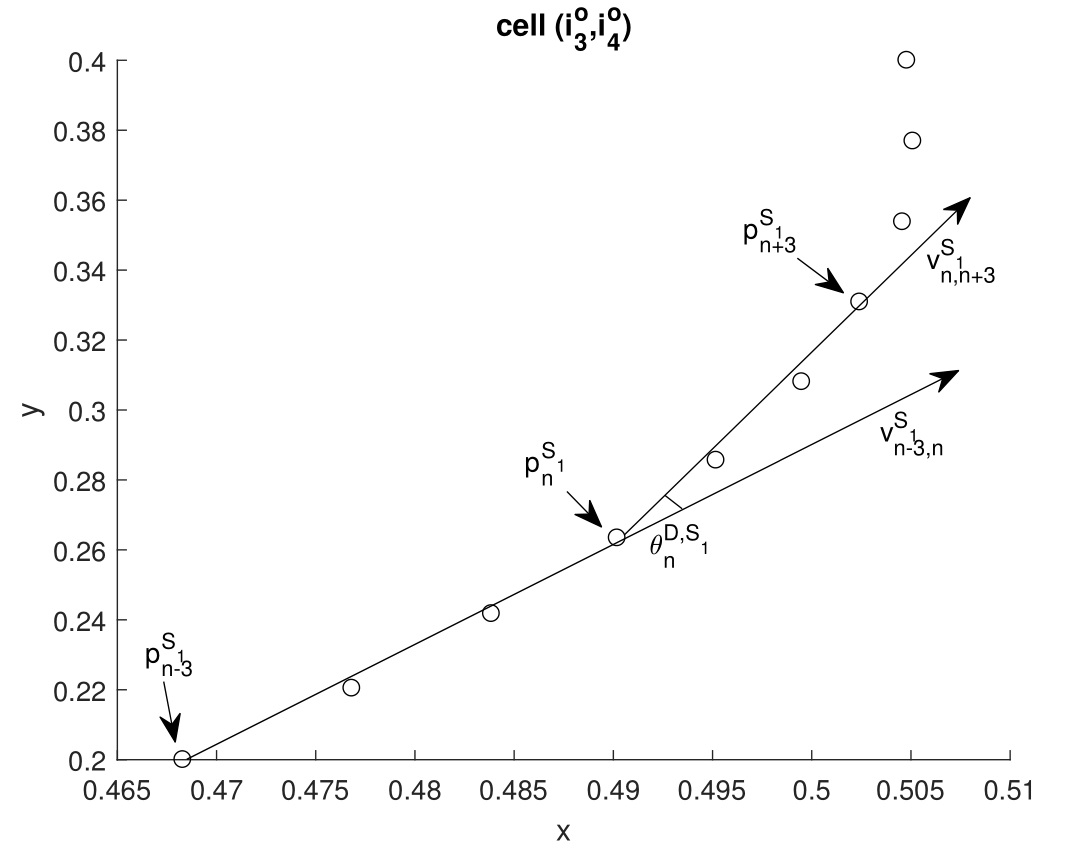}\\
(a)&(b)&(c)\\
\includegraphics[height=1.2 in]{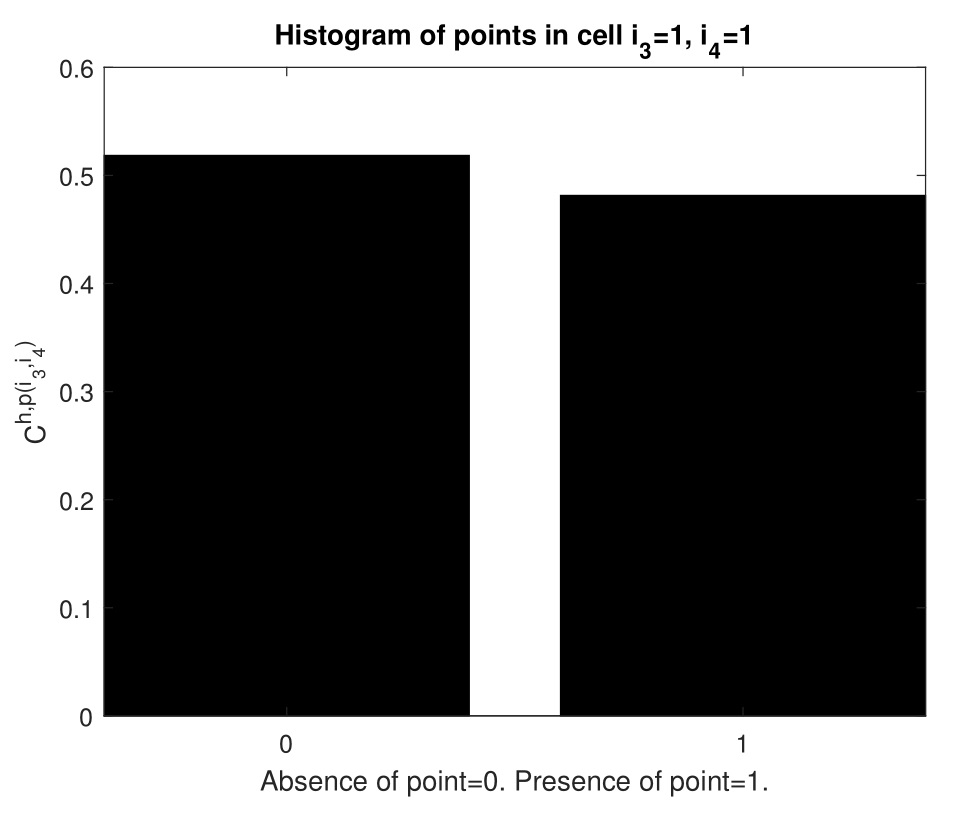}&
\includegraphics[height=1.2 in]{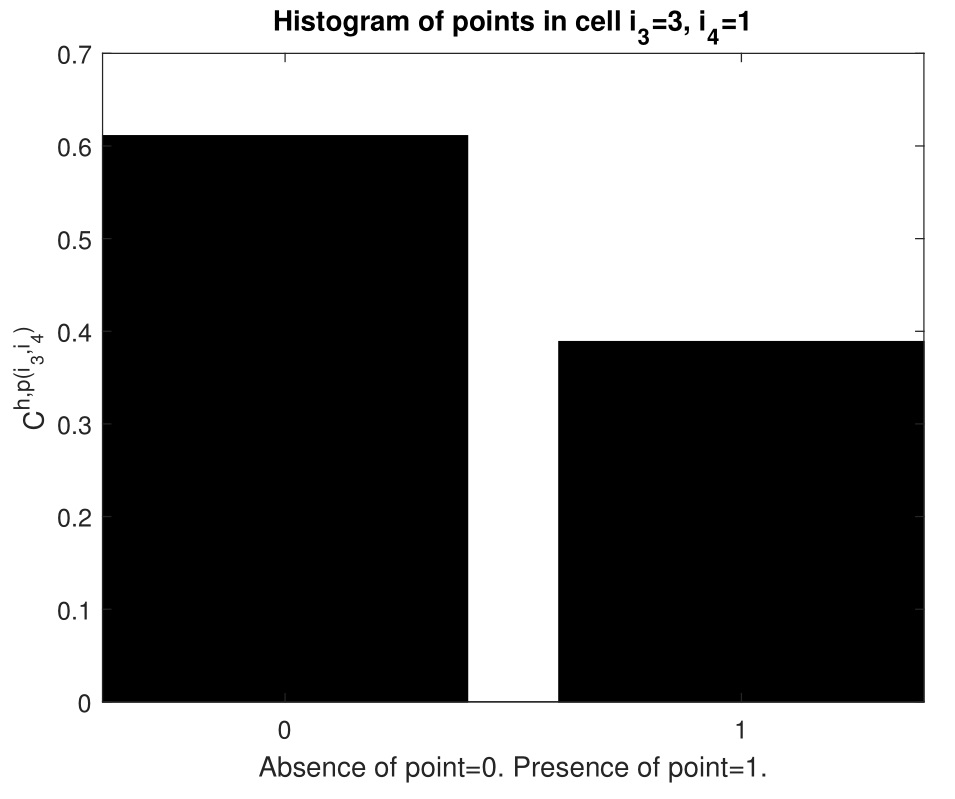}&
\includegraphics[height=1.2 in]{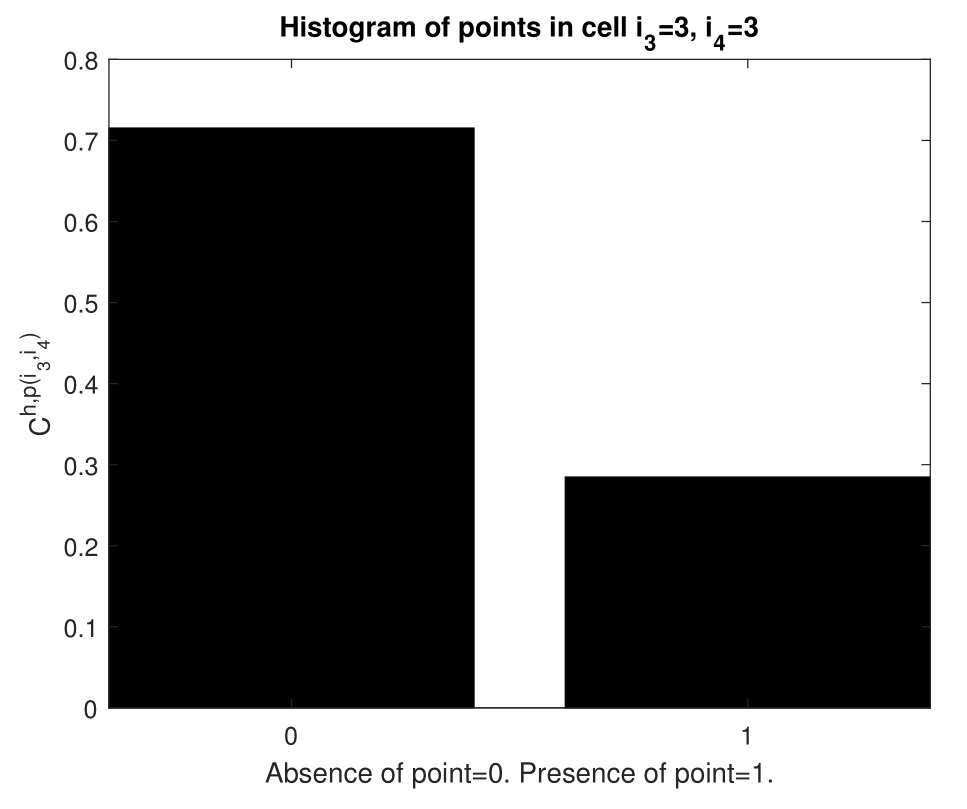}\\
(d)&(e)&(f)\\
\includegraphics[height=1.2 in]{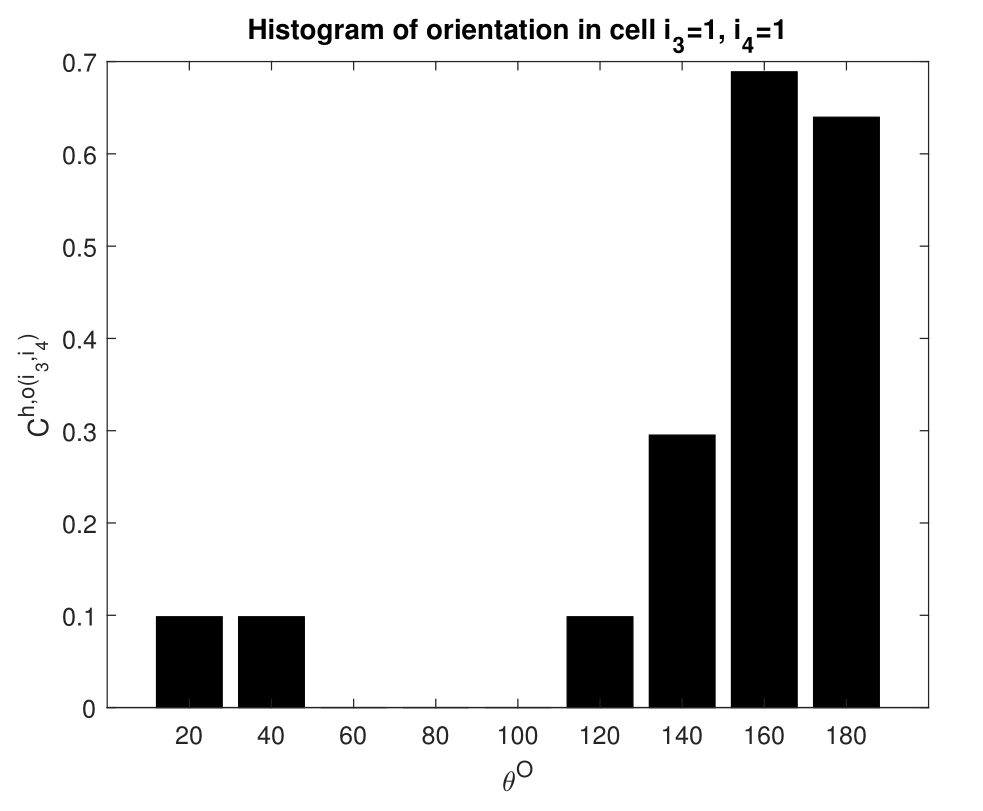}&
\includegraphics[height=1.2 in]{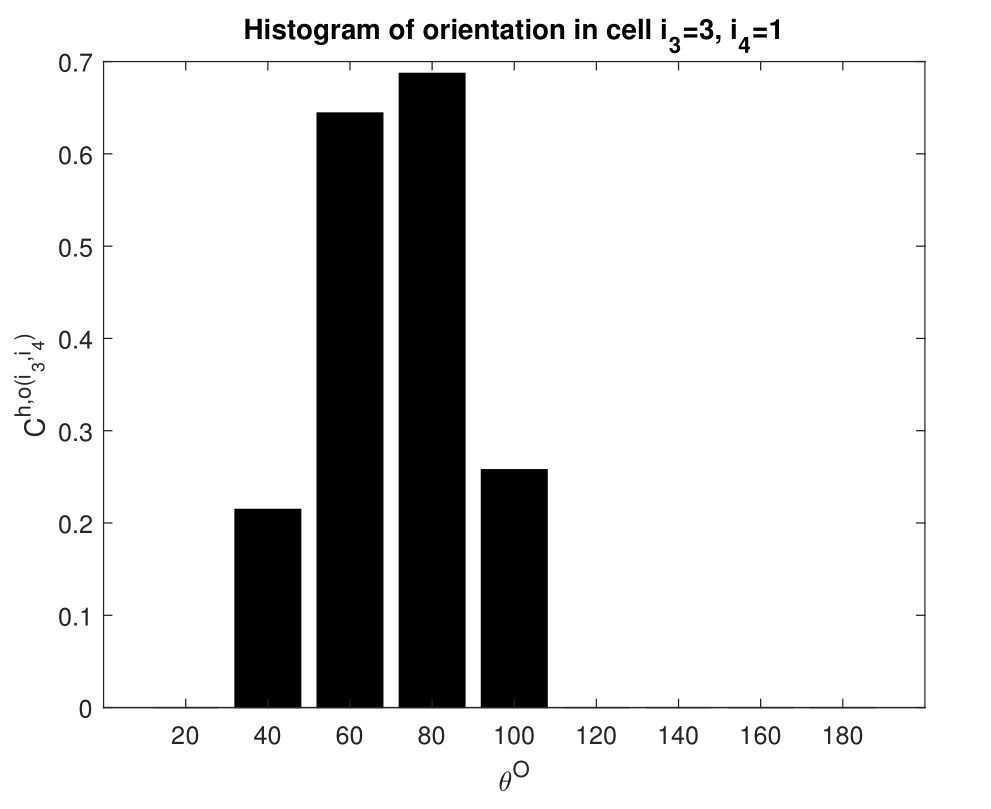}&
\includegraphics[height=1.2 in]{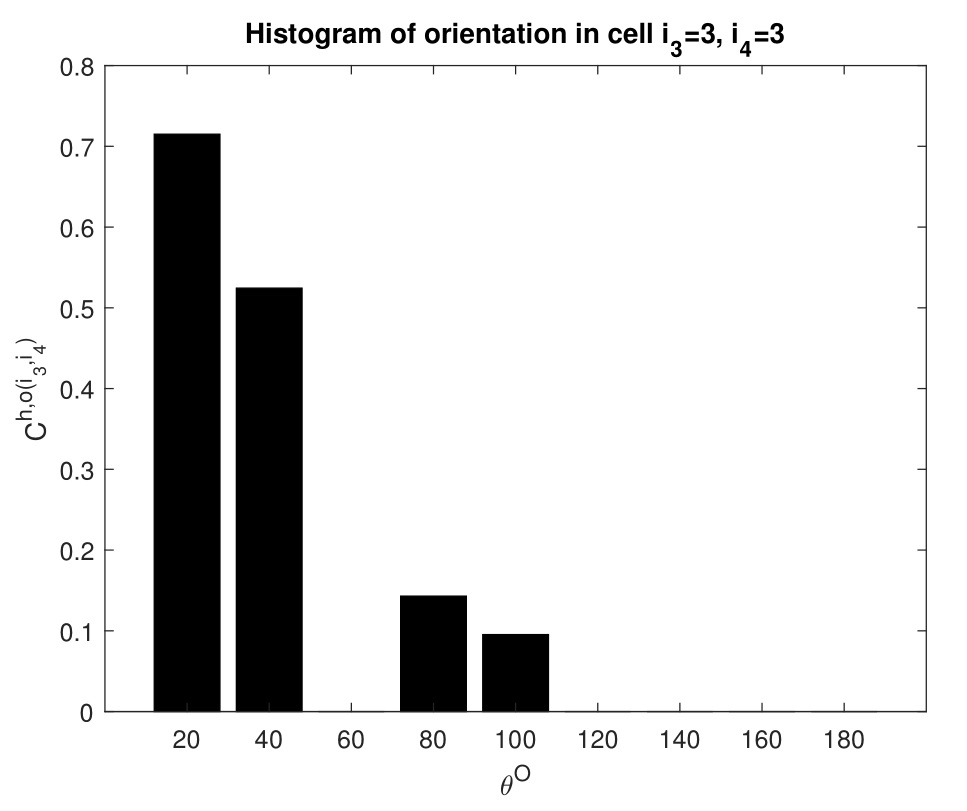}\\
(g)&(h)&(i)\\
\includegraphics[height=1.2 in]{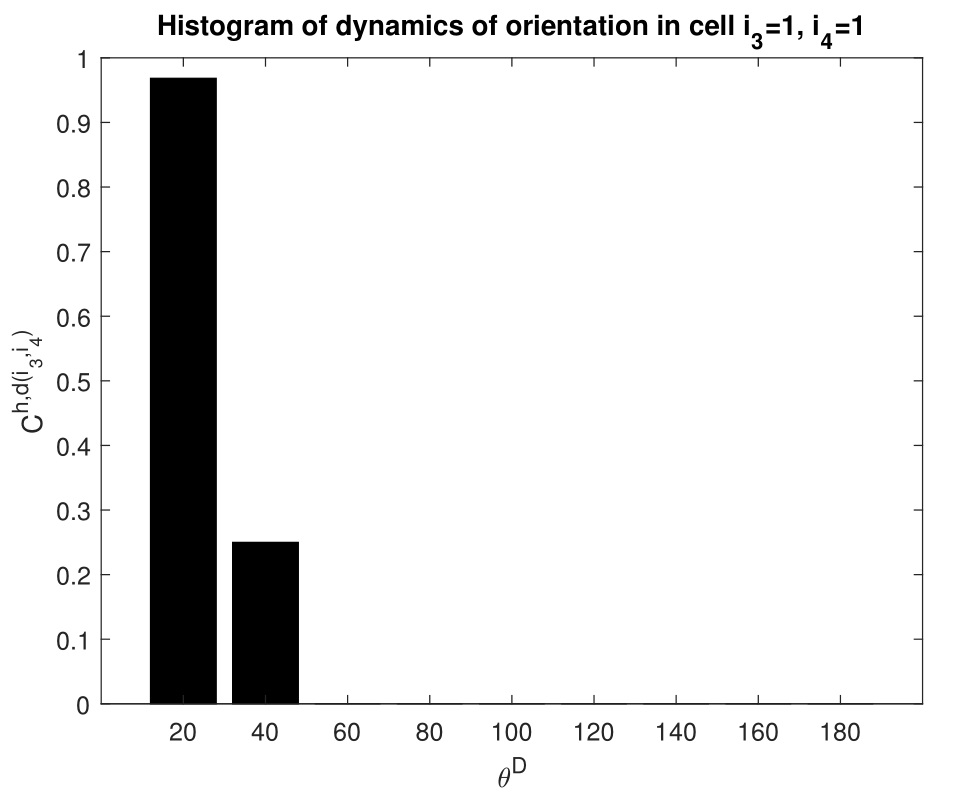}&
\includegraphics[height=1.2 in]{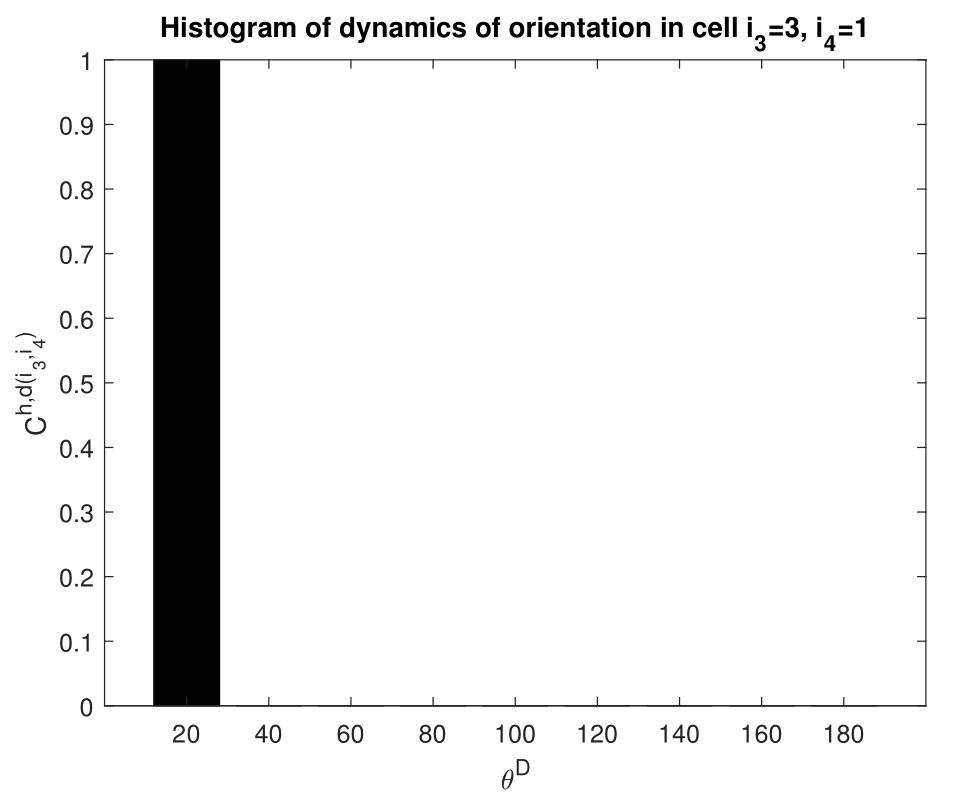}&
\includegraphics[height=1.2 in]{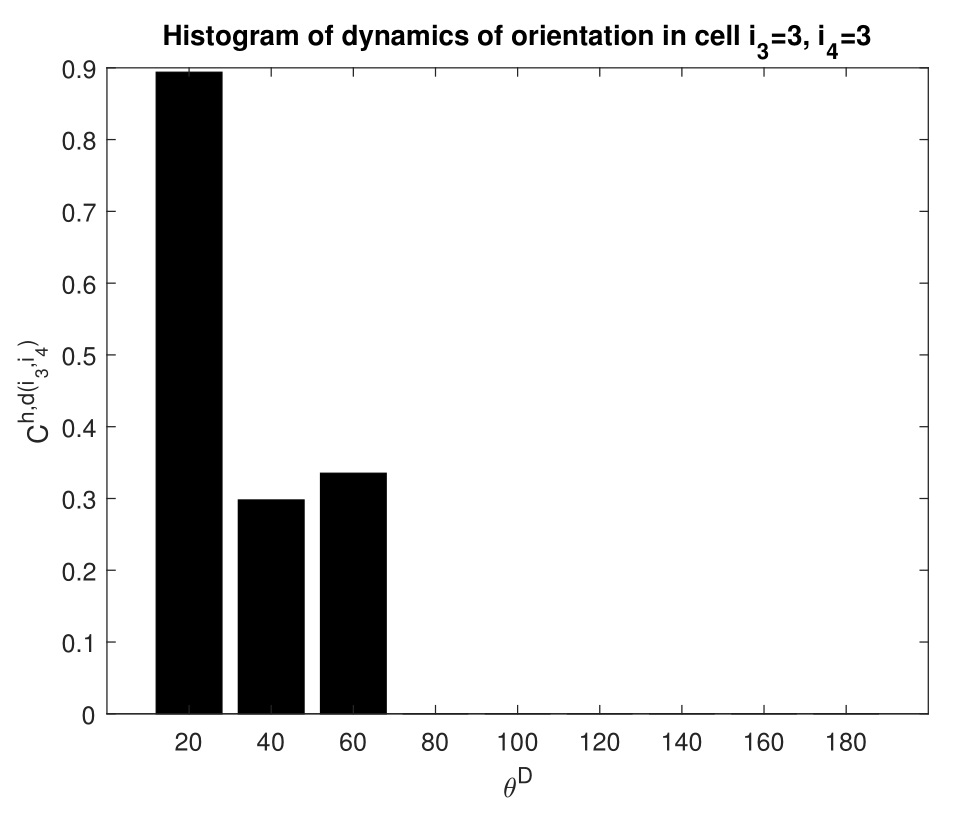}\\
(j)&(k)&(l)\\
\end{array}$
\caption{HPOD features. Orientation of stroke $\theta_n^{O,S_1}$ at the point $p_n^{S_1}$ in cell $(i_3^o,i_4^o)$ in (a) is obtained using (4) with points $p_{n-n_o}^{S_1}$ and $p_{n+n_o}^{S_1}$, $n_o=1$, as shown in (b). Dynamics of orientation of stroke $\theta_n^{D,S_1}$ at the point $p_n^{S_1}$ in cell $(i_3^o,i_4^o)$ in (a) is obtained using (5) by finding directions $v^{S_1}_{n-n_d,n}$ and $v^{S_1}_{n,n+n_d}$ using points $p_{n-n_d}^{S_1},\,\,p_{n}^{S_1},$ and $p_{n+n_d}^{S_1}$, $n_d=3$, as shown in (c). $\theta_n^{O,S_1}$ and $\theta_n^{D,S_1}$ are in the range $[0^{\circ}\,\,180^{\circ}]$ and are quantized into 9 intervals each of width $20^{\circ}$. Point $p_n^{S_1}$, orientation $\theta_n^{O,S_i}$, and dynamics of orientation $\theta_n^{D,S_i}$ are mapped spatially to location $(i_1^o,i_2^o)$ as a function of co-ordinate values of the point. The same process is applied to all the points in the character. Spatial map of the character is divided into cells indexed in x- and y-directions by $i_3$ and $i_4$, respectively, as shown in (a). Histograms of points, orientations and dynamics of orientations are computed in each of these cells. Cells $(i_3=1,\,\,i_4=1)$, $(i_3=3\,\,i_4=1)$, and $(i_3=3,\,\,i_4=3)$ are considered for showing the histograms of points, orientations and dynamics of orientations. (d)-(f) Histograms of points in cells $(i_3=1,\,\,i_4=1)$, $(i_3=3,\,\,i_4=1)$, and $(i_3=3,\,\,i_4=3)$. (g)-(i) Histograms of orientations in cells $(i_3=1,\,\,i_4=1)$, $(i_3=3,\,\,i_4=1)$, and $(i_3=3,\,\,i_4=3)$. (j)-(l) Histograms of dynamics of orientations in cells $(i_3=1,\,\,i_4=1)$, $(i_3=3,\,\,i_4=1)$, and $(i_3=3,\,\,i_4=3)$.}
\end{figure}
\end{center}

\subsubsection{Effect of stroke direction and stroke order variations on the HPOD feature representation}
\indent Let $C$, $C'$, and $C''$ be three handwritten characters and $C^M$, $C'^M$, and $C''^M$ be the corresponding character matrices. Let $(x_{-1},y_{-1})$, $(x_{o},y_{o}),$ and $(x_{1},y_{1})$ be the co-ordinate values of three arbitrary consecutive points in the characters $C$, $C'$, and $C''$ as described in Section 3.1.2. Let $(i^{-1}_1,i^{-1}_2)$, $(i^{o}_1,i^{o}_2)$, and $(i^{1}_1,i^{1}_2)$, respectively, be the indices of spatial mappings of the aforementioned points. Orientation of stroke $S_i$ at point $p_n^{S_i}$ for $n_o=1$ is \[{\theta'}_n^{O,S_i}=\tan^{-1}\left(\frac{p_{n+1}^{S_i}(2)-p_{n-1}^{S_i}(2)}{p_{n+1}^{S_i}(1)-p_{n-1}^{S_i}(1)}\right)=\tan^{-1}\left(\frac{y_1-y_{-1}}{x_1-x_{-1}}\right).\] 
Dynamics of orientation of stroke $S_i$ at point $p_n^{S_i}$ for $n_d=1$ is ${\theta'}_n^{D,S_i}=\cos^{-1}({v_{n-1,n}^{S_i}}^Tv_{n,n+1}^{S_i}).$
\[{\theta'}_n^{D,S_i}=cos^{-1}\left(\frac{(p_n^{S_i}-p_{n-1}^{S_i})^T(p_{n+1}^{S_i}-p_{n}^{S_i})}{||(p_n^{S_i}-p_{n-1}^{S_i})||_2\,||(p_{n+1}^{S_i}-p_{n}^{S_i})||_2}\right)=cos^{-1}\left(\frac{\left(\begin{array}{c}x_o-x_{-1}\\y_o-y_{-1}\end{array}\right)^T\left(\begin{array}{c}x_1-x_{o}\\y_1-y_{o}\end{array}\right)}{\Bigl\Vert\left(\begin{array}{c}x_o-x_{-1}\\y_o-y_{-1}\end{array}\right)\Bigr\Vert_2\,\,\Bigl\Vert\left(\begin{array}{c}x_1-x_{o}\\y_1-y_{o}\end{array}\right)\Bigr\Vert_2}\right).\]
Orientation of stroke $S'_i$ at point $p_{n'}^{S'_i}$ for $n_o=1$ is \[{\theta''}_{n'}^{O,S'_i}=\tan^{-1}\left(\frac{p_{n'+1}^{S'_i}(2)-p_{n'-1}^{S'_i}(2)}{p_{n'+1}^{S'_i}(1)-p_{n'-1}^{S'_i}(1)}\right)=\tan^{-1}\left(\frac{y_{-1}-y_1}{x_{-1}-x_1}\right).\]
Dynamics of orientation of stroke $S'_i$ at point $p_{n'}^{S'_i}$ for $n_d=1$ is ${\theta''}_{n'}^{D,S'_i}=\cos^{-1}({v_{n'-1,n'}^{S'_i}}^Tv_{n',n'+1}^{S'_i}).$
\[{\theta''}_{n'}^{D,S'_i}=cos^{-1}\left(\frac{(p_{n'}^{S'_i}-p_{n'-1}^{S'_i})^T(p_{n'+1}^{S'_i}-p_{n'}^{S'_i})}{||(p_{n'}^{S'_i}-p_{n'-1}^{S'_i})||_2\,\,||(p_{n'+1}^{S'_i}-p_{n'}^{S'_i})||_2}\right)=cos^{-1}\left(\frac{\left(\begin{array}{c}x_o-x_{1}\\y_o-y_{1}\end{array}\right)^T\left(\begin{array}{c}x_{-1}-x_{o}\\y_{-1}-y_{o}\end{array}\right)}{\Bigl\Vert\left(\begin{array}{c}x_o-x_{1}\\y_o-y_{1}\end{array}\right)\Bigr\Vert_2\,\Bigl\Vert\left(\begin{array}{c}x_{-1}-x_{o}\\y_{-1}-y_{o}\end{array}\right)\Bigr\Vert_2}\right).\]
Orientation of stroke $S''_{i+1}$ at point $p_{n}^{S''_{i+1}}$ for $n_o=1$ is \[{\theta'''}_{n}^{O,S''_{i+1}}=\tan^{-1}\left(\frac{p_{n+1}^{S''_{i+1}}(2)-p_{n-1}^{S''_{i+1}}(2)}{p_{n+1}^{S''_{i+1}}(1)-p_{n-1}^{S''_{i+1}}(1)}\right)=\tan^{-1}\left(\frac{y_{1}-y_{-1}}{x_{1}-x_{-1}}\right).\]
Dynamics of orientation of stroke $S''_{i+1}$ at point $p_{n}^{S''_{i+1}}$ for $n_d=1$ is ${\theta'''}_{n}^{D,S''_{i+1}}=\cos^{-1}({v_{n-1,n}^{S''_{i+1}}}^Tv_{n,n+1}^{S''_{i+1}}).$
\[{\theta'''}_{n}^{D,S''_{i+1}}=cos^{-1}\left(\frac{(p_{n}^{S''_{i+1}}-p_{n-1}^{S''_{i+1}})^T(p_{n+1}^{S''_{i+1}}-p_{n}^{S''_{i+1}})}{||(p_{n}^{S''_{i+1}}-p_{n-1}^{S''_{i+1}})||_2\,||(p_{n+1}^{S''_{i+1}}-p_{n}^{S''_{i+1}})||_2}\right)=cos^{-1}\left(\frac{\left(\begin{array}{c}x_o-x_{-1}\\y_o-y_{-1}\end{array}\right)^T\left(\begin{array}{c}x_{1}-x_{o}\\y_{1}-y_{o}\end{array}\right)}{\Bigl\Vert\left(\begin{array}{c}x_o-x_{-1}\\y_o-y_{-1}\end{array}\right)\Bigr\Vert_2\,\Bigl\Vert\left(\begin{array}{c}x_{1}-x_{o}\\y_{1}-y_{o}\end{array}\right)\Bigr\Vert_2}\right).\] 
\indent Spatial representation of a point is independent of stroke direction and stroke order variations as observed in section 3.5.2. The orientations of strokes at points $p_n^{S_i}$, $p_{n'}^{S'_i}$, and $p_n^{S''_{i+1}}$ are ${\theta'}_{n}^{O,S_{i}},\,\,\,{\theta''}_{n'}^{O,S'_{i}},$ and ${\theta'''}_{n}^{O,S''_{i+1}}$, respectively, and are identical. The dynamics of orientations of strokes at points $p_n^{S_i}$, $p_{n'}^{S'_i}$, and $p_n^{S''_{i+1}}$ are ${\theta'}_{n}^{D,S_{i}},\,\,\,{\theta''}_{n'}^{D,S'_{i}},$ and ${\theta'''}_{n}^{D,S''_{i+1}}$, respectively, and are also identical. So the variations in stroke direction or stroke order or both do not change these values and the same applies to all the points in all the characters. Orientation of stroke and dynamics of orientation of stroke at a point are mapped spatially as a function of co-ordinate values of that point and so are independent of stroke direction and stroke order variations and the same applies to all the points in the character. HPOD features are obtained from spatial mapping of points, orientations and dynamics of orientations and so are independent of variations in order and direction of strokes in characters.

\section{Experiments and results}
\subsection{Support vector machines (SVM)}
\indent Support vector machines (SVM) classifier as described in Cortes et al. \cite{clsfff} is chosen for evaluating the character discriminative capability of different features described above. SVM is a two class classifier and has good generalization capability with good recognition accuracy even when trained with dataset of small size. SVM classifier uses regularization parameter $\beta$ to control its capacity and kernel function $k^r(X,Y)$ to implicitly non-linearly map feature vectors to a high dimensional space. Since there are $N_{ct}$ classes, one-versus-one approach is used as given in Platt et al. \cite{daga} by considering $\frac{N_{ct}(N_{ct}-1)}{2}$ two class SVM classifiers for classification of character samples. Let $(w^{i,j},w^{i,j}_0),\,\,\,w^{i,j}=w^{i,j}_1,\dots,w^{i,j}_{N_{ftr}},\,\,\,1\le i<j\le N_{ct},$ be the parameters defining $\frac{N_{ct}(N_{ct}-1)}{2}$ two class classifiers trained by considering pair of data sets $(d^{tr}_i,\,d^{tr}_j),\,\,\,1\le i<j\le N_{ct}$. $N_{ftr}$ is the dimension of a feature vector. The final decision of this collection of two class classifiers on class label of a feature vector $X^{ts}_{m_1}$ is made through $N_{ct}-1$ intermediate decisions. Starting from $k=1,\,j=2$, the classifier $(w^{k,j},w^{k,j}_0)$ is chosen to classify the feature vector $X^{ts}_{m_1}$.
\[\mbox{If}\,\,\, ({w^{k,j}}^T\,X^{ts}_{m_1}+w^{k,j}_0)>0,\,\,\,k=k,\,\,\,j=\max(k,j)+1,\,\,\,j\le N_{ct},\]
\[\mbox{if}\,\,\, ({w^{k,j}}^T\,X^{ts}_{m_1}+w^{k,j}_0)\le0,\,\,\,k=j,\,\,\,j=\max(k,j)+1,\,\,\,j\le N_{ct}.\]
At the end of $N_{ct}-1$ such decision steps the value of $k$ is the decision of the collection of two class classifiers.\\
The accuracy $A_c$ of SVM classifier is 
$A_c=\frac{\mbox{total number of correct classifications}}{\mbox{total number of classifications}}.$\\[2.5pt]
Parameters of SVM classifiers for different features have been tuned so as to make the SVMs produce the best classification performance on the test dataset. Parameters of the classifiers for different features are given below.\\
$k^r(X,Y)=\exp(-\frac{1}{\Upsilon^2}||X-Y||^2)$.\\ 
$\beta=1024$ and $\Upsilon=10$ for ST features\\
$\beta=1024$ and $\Upsilon=28$ for DFT features\\
$\beta=1024$ and $\Upsilon=28$ for DCT features\\
$\beta=1024$ and $\Upsilon=20$ for DWT features\\
$\beta=1024$ and $\Upsilon=10$ for SP features\\
$\beta=1024$ and $\Upsilon=10$ for HOG features\\
$\beta=1024$ and $\Upsilon=10$ for HPOD features\\

\subsection{Feature parameters}
\indent The values of the different feature parameters are given below.\\ 
$N_{ftr}$ is the dimension of a feature vector.\\
ST feature parameters: $N_{ftr}=N_{ST}+2=258$.\\
DFT feature parameters: $N_{ftr}=N_{DFT}+2=258$.\\
DCT feature parameters: $N_{ftr}=N_{DCT}+2=258$.\\
DWT feature parameters: $N_{ftr}=N_{DWT}+2=258$, filter length $=2$, filter type=`Daubechies'.\\
SP feature parameters: $\Delta^{sp}=0.0357,\,\,\,N_{ftr}=N_{SP}+2=786$.\\
HOG feature parameters: $\Delta^{sp}=0.0278,\,\,\,N_{cl}=6,\,\,\,\Delta^o=20,\,\,\,\Delta^b=1,\,\,\,\Delta^{ov}=0,\,\,\,N_{ftr}=N_{HOG}+2=326$.\\
HPOD feature parameters: $\Delta^{sp}=0.0278,\,\,\, n_o=1,\,\,\,n_d=3,\,\,\,\Delta^{\theta_o}=20,\,\,\,\Delta^{\theta_d}=20,\,\,\,\Delta^{pcl}=6,\,\,\,\Delta^{ocl}=6,\,\,\,\Delta^{dcl}=6,\,\,\,\Delta_p^{ov}=3,\,\,\,\Delta_o^{ov}=3,\,\,\,\Delta_d^{ov}=3,\,\,\,N_{ftr}=N_{HPOD}+2=722$.\\

\subsection{Classifier performance on different features}
\begin{table}[ht!]
\caption{Classification accuracies(\%) of SVM classifiers on training set and test set of different feature vectors with different dimensions are given. Number of character classes is 96. Average number of samples per class in training set is 133 and in testing set is 29.}
\begin{center}\begin{tabular}[ht]{|c|c|c|c|c|c|c|c|}
\hline
 &ST&DFT&DCT&DWT&SP&HOG&HPOD\\
\hline
Dimension&258&258&258&258&786&326&722\\
\hline
Training set accuracies&99.9&100&98.2&99.6&100&99.8&100\\
\hline
Test set accuracies&89.2&90.2&86.7&88.3&76.7&77.6&92.9\\
\hline
\end{tabular}\end{center}\end{table}
    
\section{Conclusion}
\indent Features used in the studies on Hindi online handwritten character recognition are either dependent on variations in order and direction of strokes in characters or do not have enough character discriminative capability. Features in these studies have been extracted from different datasets or character set of different sizes making it difficult to compare performance of classifiers trained with these features.\\
\indent HPOD features are developed that are independent of variations in order and direction of strokes in characters. The method for extraction of HPOD features maps features like co-ordinates of points, orientations of strokes at points and dynamics of orientations of strokes at points spatially as a function of co-ordinate values of the points and computes the histograms of these features from different regions in the spatial map. \\
\indent Features like ST, DFT, DCT, DWT, SP, and HOG used in other studies are considered in this study for comparison of their discriminative properties with that of the HPOD features developed in this study. ST, DFT, DCT, and DWT features are dependent on variations in order and direction of strokes in characters whereas SP and HOG features are independent of such variations.\\
\indent The discriminative capabilities of different features are measured in terms of the classification accuracy of SVM classifiers trained independently with these features extracted from the same training set. The classification performance of these classifiers are tested on the same testing set. SVM trained with HPOD features has the highest accuracy of 92.9\% compared to the SVM classifiers trained with the other features. It shows that character discriminative capability of HPOD features is better than the other features considered in this study.



\begin{thebibliography}{24}


\ifx \showCODEN    \undefined \def \showCODEN     #1{\unskip}     \fi
\ifx \showDOI      \undefined \def \showDOI       #1{#1}\fi
\ifx \showISBNx    \undefined \def \showISBNx     #1{\unskip}     \fi
\ifx \showISBNxiii \undefined \def \showISBNxiii  #1{\unskip}     \fi
\ifx \showISSN     \undefined \def \showISSN      #1{\unskip}     \fi
\ifx \showLCCN     \undefined \def \showLCCN      #1{\unskip}     \fi
\ifx \shownote     \undefined \def \shownote      #1{#1}          \fi
\ifx \showarticletitle \undefined \def \showarticletitle #1{#1}   \fi
\ifx \showURL      \undefined \def \showURL       {\relax}        \fi
\providecommand\bibfield[2]{#2}
\providecommand\bibinfo[2]{#2}
\providecommand\natexlab[1]{#1}
\providecommand\showeprint[2][]{arXiv:#2}

\bibitem[Abbate et~al\mbox{.}(2002)]%
        {ak}
\bibfield{author}{\bibinfo{person}{A. Abbate}, \bibinfo{person}{C.~M.
  DeCusatis}, {and} \bibinfo{person}{P.~K. Das}.}
  \bibinfo{year}{2002}\natexlab{}.
\newblock \bibinfo{booktitle}{\emph{Wavelets and Subbands: Fundamentals and
  Applications}}.
\newblock \bibinfo{publisher}{Birkhauser}.
\newblock


\bibitem[Bahlmann(2006)]%
        {ftra}
\bibfield{author}{\bibinfo{person}{C. Bahlmann}.}
  \bibinfo{year}{2006}\natexlab{}.
\newblock \showarticletitle{Directional features in online handwriting
  recognition}.
\newblock \bibinfo{journal}{\emph{Pattern Recogn.}} \bibinfo{volume}{39},
  \bibinfo{number}{1} (\bibinfo{year}{2006}), \bibinfo{pages}{115--125}.
\newblock


\bibitem[Belhe et~al\mbox{.}(2012)]%
        {troaw}
\bibfield{author}{\bibinfo{person}{S. Belhe}, \bibinfo{person}{C. Paulzagade},
  \bibinfo{person}{A. Deshmukh}, \bibinfo{person}{S. Jetley}, {and}
  \bibinfo{person}{K. Mehrotra}.} \bibinfo{year}{2012}\natexlab{}.
\newblock \showarticletitle{Hindi handwritten word recognition using HMM and
  symbol tree}. In \bibinfo{booktitle}{\emph{Proceeding of the Workshop on
  Document Analysis and Recognition}} \emph{(\bibinfo{series}{DAR `12})}.
  \bibinfo{pages}{9--14}.
\newblock


\bibitem[Bhattacharya et~al\mbox{.}(2007)]%
        {ftrh}
\bibfield{author}{\bibinfo{person}{U. Bhattacharya}, \bibinfo{person}{B.~K.
  Gupta}, {and} \bibinfo{person}{S.~K. Parui}.}
  \bibinfo{year}{2007}\natexlab{}.
\newblock \showarticletitle{Direction code based features for recognition of
  online handwritten characters of Bangla}. In
  \bibinfo{booktitle}{\emph{Proceedings of the 9th International Conference on
  Document Analysis and Recognition}} \emph{(\bibinfo{series}{ICDAR `07})}.
\newblock


\bibitem[Connell et~al\mbox{.}(2000)]%
        {troan}
\bibfield{author}{\bibinfo{person}{S.~D. Connell}, \bibinfo{person}{R.~M.~K.
  Sinha}, {and} \bibinfo{person}{A.~K. Jain}.} \bibinfo{year}{2000}\natexlab{}.
\newblock \showarticletitle{Recognition of unconstrained on-line Devanagari
  characters}. In \bibinfo{booktitle}{\emph{Proceedings of 15th International
  Conference on Pattern Recognition}} \emph{(\bibinfo{series}{ICPR 2000})}.
\newblock


\bibitem[Cortes and Vapnik(1995)]%
        {clsfff}
\bibfield{author}{\bibinfo{person}{C. Cortes} {and} \bibinfo{person}{V.
  Vapnik}.} \bibinfo{year}{1995}\natexlab{}.
\newblock \showarticletitle{Support-vector networks}.
\newblock \bibinfo{journal}{\emph{Mach. Learn.}} \bibinfo{volume}{20},
  \bibinfo{number}{3} (\bibinfo{year}{1995}).
\newblock


\bibitem[Dalal and Triggs(2005)]%
        {ftrm}
\bibfield{author}{\bibinfo{person}{N. Dalal} {and} \bibinfo{person}{B.
  Triggs}.} \bibinfo{year}{2005}\natexlab{}.
\newblock \showarticletitle{Histograms of oriented gradients for human
  detection}. In \bibinfo{booktitle}{\emph{Proceedings of the 2005 IEEE
  Computer Society Conference on Computer Vision and Pattern Recognition}}
  \emph{(\bibinfo{series}{CVPR `05})}. \bibinfo{pages}{886--893}.
\newblock


\bibitem[Duda et~al\mbox{.}(2000)]%
        {troam}
\bibfield{author}{\bibinfo{person}{R.~O. Duda}, \bibinfo{person}{P.~E. Hart},
  {and} \bibinfo{person}{D.~G. Stork}.} \bibinfo{year}{2000}\natexlab{}.
\newblock \bibinfo{booktitle}{\emph{Pattern Classification (2nd edition)}}.
\newblock \bibinfo{publisher}{Wiley}, \bibinfo{address}{New York, NY, USA}.
\newblock


\bibitem[Hamanaka et~al\mbox{.}(1993)]%
        {ftre}
\bibfield{author}{\bibinfo{person}{M. Hamanaka}, \bibinfo{person}{K. Yamada},
  {and} \bibinfo{person}{J. Tsukumo}.} \bibinfo{year}{1993}\natexlab{}.
\newblock \showarticletitle{On-line Japanese character recognition experiments
  by an off-line method based on normalization-cooperated feature extraction}.
  In \bibinfo{booktitle}{\emph{Proceedings of 2nd International Conference on
  Document Analysis and Recognition}} \emph{(\bibinfo{series}{ICDAR `93})}.
\newblock


\bibitem[Jaeger et~al\mbox{.}(2001)]%
        {ftrl}
\bibfield{author}{\bibinfo{person}{S. Jaeger}, \bibinfo{person}{S. Manke},
  \bibinfo{person}{J. Reichert}, {and} \bibinfo{person}{A. Waibel}.}
  \bibinfo{year}{2001}\natexlab{}.
\newblock \showarticletitle{Online handwriting recognition: the NPen++
  recognizer}.
\newblock \bibinfo{journal}{\emph{Int. J. Doc. Aanal. Recognit.}}
  \bibinfo{volume}{3}, \bibinfo{number}{3} (\bibinfo{year}{2001}),
  \bibinfo{pages}{169--180}.
\newblock


\bibitem[Kawamura et~al\mbox{.}(1992)]%
        {ftrg}
\bibfield{author}{\bibinfo{person}{A. Kawamura}, \bibinfo{person}{K. Yura},
  \bibinfo{person}{T. Hayama}, \bibinfo{person}{Y. Hidai}, \bibinfo{person}{T.
  Minamikawa}, \bibinfo{person}{A. Tanaka}, {and} \bibinfo{person}{S. Masuda}.}
  \bibinfo{year}{1992}\natexlab{}.
\newblock \showarticletitle{On-line recognition of freely handwritten Japanese
  characters using directional feature densities}. In
  \bibinfo{booktitle}{\emph{Proceedings of 11th International Conference on
  Pattern Recognition}} \emph{(\bibinfo{series}{ICPR `92})}.
\newblock


\bibitem[Kubatur et~al\mbox{.}(2012)]%
        {troav}
\bibfield{author}{\bibinfo{person}{S. Kubatur}, \bibinfo{person}{M.~S. Ahmed},
  {and} \bibinfo{person}{M. Ahmadi}.} \bibinfo{year}{2012}\natexlab{}.
\newblock \showarticletitle{A neural network approach to online Devanagari
  handwritten character recognition}. In
  \bibinfo{booktitle}{\emph{International Conference on High Performance
  Computing \& Simulation}} \emph{(\bibinfo{series}{HPCS `12})}.
\newblock


\bibitem[Kunte and Samuel(2000)]%
        {ftri}
\bibfield{author}{\bibinfo{person}{S.~R. Kunte} {and} \bibinfo{person}{S.
  Samuel}.} \bibinfo{year}{2000}\natexlab{}.
\newblock \showarticletitle{Wavelet features based online recognition of
  handwritten Kannada characters}. In \bibinfo{booktitle}{\emph{Journal
  Visualization Society of Japan}}.
\newblock


\bibitem[Ma et~al\mbox{.}(2009)]%
        {ftrj}
\bibfield{author}{\bibinfo{person}{L. Ma}, \bibinfo{person}{Q. Huo}, {and}
  \bibinfo{person}{Y. Shi}.} \bibinfo{year}{2009}\natexlab{}.
\newblock \showarticletitle{A study of feature design for online handwritten
  Chinese character recognition based on continuous-density hidden Markov
  models}. In \bibinfo{booktitle}{\emph{10th International Conference on
  Document Analysis and Recognition}} \emph{(\bibinfo{series}{ICDAR `09})}.
\newblock


\bibitem[Mehrotra et~al\mbox{.}(2013)]%
        {troax}
\bibfield{author}{\bibinfo{person}{K. Mehrotra}, \bibinfo{person}{S. Jetley},
  \bibinfo{person}{A. Deshmukh}, {and} \bibinfo{person}{S. Belhe}.}
  \bibinfo{year}{2013}\natexlab{}.
\newblock \showarticletitle{Unconstrained handwritten Devanagari character
  recognition using convolutional neural networks}. In
  \bibinfo{booktitle}{\emph{Proceedings of the 4th International Workshop on
  Multilingual OCR}} \emph{(\bibinfo{series}{MOCR `13})}.
  \bibinfo{pages}{15:1--15:5}.
\newblock


\bibitem[of~Indian Standard~(BIS)(1991)]%
        {troiscii}
\bibfield{author}{\bibinfo{person}{Bureau of Indian Standard~(BIS)}.}
  \bibinfo{year}{1991}\natexlab{}.
\newblock \bibinfo{title}{Indian script code for information interhange
  (ISCII)}.
\newblock
\newblock


\bibitem[Okamoto and Yamamoto(1999a)]%
        {ftrk}
\bibfield{author}{\bibinfo{person}{M. Okamoto} {and} \bibinfo{person}{K.
  Yamamoto}.} \bibinfo{year}{1999}\natexlab{a}.
\newblock \showarticletitle{On-line handwriting character recognition using
  direction-change features that consider imaginary strokes}.
\newblock \bibinfo{journal}{\emph{Pattern Recogn.}} \bibinfo{volume}{32},
  \bibinfo{number}{7} (\bibinfo{year}{1999}), \bibinfo{pages}{1115--1128}.
\newblock


\bibitem[Okamoto and Yamamoto(1999b)]%
        {ftrc}
\bibfield{author}{\bibinfo{person}{M. Okamoto} {and} \bibinfo{person}{K.
  Yamamoto}.} \bibinfo{year}{1999}\natexlab{b}.
\newblock \showarticletitle{On-line handwritten character recognition method
  using directional features and clockwise/counterclockwise direction-change
  features}. In \bibinfo{booktitle}{\emph{Proceedings of the Fifth
  International Conference on Document Analysis and Recognition}}
  \emph{(\bibinfo{series}{ICDAR `99})}.
\newblock


\bibitem[Oppenheim et~al\mbox{.}(2006)]%
        {al}
\bibfield{author}{\bibinfo{person}{A.~V. Oppenheim}, \bibinfo{person}{R~W.
  Schafer}, {and} \bibinfo{person}{J.~R. Buck}.}
  \bibinfo{year}{2006}\natexlab{}.
\newblock \bibinfo{booktitle}{\emph{Discrete-Time Signal Processing}}.
\newblock \bibinfo{publisher}{Dorling Kindersley (India)}.
\newblock


\bibitem[Platt et~al\mbox{.}(2000)]%
        {daga}
\bibfield{author}{\bibinfo{person}{J. Platt}, \bibinfo{person}{N. Cristianini},
  {and} \bibinfo{person}{J. Shawe-Taylor}.} \bibinfo{year}{2000}\natexlab{}.
\newblock \showarticletitle{Large margin DAGs for multiclass classification}.
  In \bibinfo{booktitle}{\emph{Advances in Neural Information Processing
  Systems}}.
\newblock


\bibitem[Sundaram and Ramakrishnan(2013)]%
        {troajj}
\bibfield{author}{\bibinfo{person}{S. Sundaram} {and} \bibinfo{person}{A.~G.
  Ramakrishnan}.} \bibinfo{year}{2013}\natexlab{}.
\newblock \showarticletitle{Attention-feedback based robust segmentation of
  online handwritten isolated Tamil words}.
\newblock \bibinfo{journal}{\emph{ACM Transactions on Asian Language
  Information Processing (TALIP)}} \bibinfo{volume}{12}, \bibinfo{number}{1}
  (\bibinfo{year}{2013}), \bibinfo{pages}{4:1--4:25}.
\newblock


\bibitem[Swethalakshmi(2008)]%
        {ftrmm}
\bibfield{author}{\bibinfo{person}{H. Swethalakshmi}.}
  \bibinfo{year}{2008}\natexlab{}.
\newblock \emph{\bibinfo{title}{Online handwritten character recognition for
  Devanagari and Tamil scripts using support vector machines}}.
\newblock \bibinfo{thesistype}{Master's\ thesis}. \bibinfo{school}{Indian
  Institute of Technology, Madras}.
\newblock


\bibitem[Tappert et~al\mbox{.}(1990)]%
        {troaa}
\bibfield{author}{\bibinfo{person}{C. Tappert}, \bibinfo{person}{C. Suen},
  {and} \bibinfo{person}{T. Wakahara}.} \bibinfo{year}{1990}\natexlab{}.
\newblock \showarticletitle{State of the art in online handwriting
  recognition}.
\newblock \bibinfo{journal}{\emph{IEEE Trans. Pattern Anal. Mach. Intell.}}
  \bibinfo{volume}{12}, \bibinfo{number}{8} (\bibinfo{year}{1990}),
  \bibinfo{pages}{787--808}.
\newblock


\bibitem[toolkit(2014)]%
        {trohp}
\bibfield{author}{\bibinfo{person}{Lipi toolkit}.}
  \bibinfo{year}{2014}\natexlab{}.
\newblock \bibinfo{booktitle}{\emph{HP labs India Indic handwriting datasets}}.
\newblock
\urldef\tempurl%
\url{http://lipitk.sourceforge.net/datasets/ dvngchardata.htm}
\showURL{%
\tempurl}


\end{thebibliography}
\end{document}